\documentclass{article} %
\usepackage{nips12submit_e,times}
\usepackage{graphicx}
\usepackage{natbib}
\usepackage{amsmath,amssymb} %
\usepackage{color}
\usepackage[small]{caption}
\usepackage{float}
\usepackage{subfigure}
\title{Learnable Pooling Regions for Image Classification}

\author{
Mateusz Malinowski \\
Computer Vision and Multimodal Computing\\
Max Planck Institute for Informatics\\
Campus E1 4, 66123 Saarbr\"{u}cken, Germany \\
\texttt{mmalinow at mpi-inf.mpg.de} \\
\And
Mario Fritz \\
Computer Vision and Multimodal Computing\\
Max Planck Institute for Informatics\\
Campus E1 4, 66123 Saarbr\"{u}cken, Germany \\
\texttt{mfritz at mpi-inf.mpg.de} \\
}

\nipsfinalcopy %

\newcommand{\pnorm}[2]{\ensuremath{|| #1 ||_{#2}}}

\newcommand{\twoNorm}[1]{\pnorm{#1}{l_2}}

\newcommand{\bs}[1]{\ensuremath{\boldsymbol{#1}}}

\newcommand{\bestTwo}[2]{\ensuremath{#1^{*,#2}}}

\DeclareMathOperator* {\minimize}{minimize}

\begin{document} 
	
\maketitle 

\begin{abstract}
Biologically inspired, from the early HMAX model to Spatial Pyramid Matching, pooling has played an important role in visual recognition pipelines. Spatial pooling, by grouping of local codes, equips these methods with a certain degree of robustness to translation and deformation yet preserving important spatial information.
Despite the predominance of this approach in current recognition systems, we have seen  little progress to fully adapt  the pooling strategy to the task at hand. This paper proposes a model for learning task dependent pooling scheme -- including previously proposed hand-crafted pooling schemes as a particular instantiation.
In our work, we investigate the role of different regularization terms showing that the smooth regularization term is crucial to achieve strong performance using the presented architecture.
Finally, we propose an efficient and parallel method to train the model.
Our experiments show improved performance over hand-crafted pooling schemes on the CIFAR-10 and CIFAR-100 datasets -- in particular improving the state-of-the-art to $56.29\%$ on the latter.
\end{abstract}

\section{Introduction}

Spatial pooling plays a crucial role in modern object recognition and detection systems. Motivated from biology \citep{Riesenhuber:1999:HMAX} and statistics of locally orderless images \citep{orderless}, the spatial pooling approach has been found useful as an intermediate step of many today's computer vision methods. For instance, the most popular visual descriptors such as SIFT \citep{Lowe:2004:SIFT} and HOG \citep{Dalal:2005:HOG}, which compute local histograms of gradients, can be in fact seen as a special version of the spatial pooling strategy. In order to form more robust features under translation or small object deformations, activations of the codes are  pooled over larger areas in a spatial pyramid scheme \citep{lazebnik2006beyond,yang2009linear}.
Unfortunately, this critical decision, namely the spatial division, is most prominently based on hand-crafted algorithms and therefore data independent.

\paragraph{Related Work}
As large amounts of training data is available to us today,, there is an increasing interest to push the boundary of learning based approaches towards fully optimized and adaptive architectures where design choices, that would potentially constrain or bias a model, are kept to a minimum. 
Neural networks have a great tradition of approaching hierarchical learning problems and training intermediate representations
\citep{ranzato07cvpr,cat12icml}.
Along this line, we propose a learnable spatial pooling strategy that can shape the pooling regions in a discriminative manner. Our architecture has a direct interpretation as a pooling strategy and therefore subsumes popular spatial pyramids as a special case.
Yet we have the freedom to investigate different regularization terms that lead to new pooling strategies when optimized jointly with the classifier.

Recent progress has been made in learning pooling regions in the context of image classification using the Spatial Pyramid Matching (SPM)  pipeline \citep{lazebnik2006beyond,yang2009linear}.  
\citet{yangqing11nips}, \citet{jia2012beyond} and \citet{feng11cvpr} have investigated how to further liberate the recognition from preconceptions of the hand crafted recognition pipelines, and include the pooling strategy into the optimization framework jointly with the classifier. However, these methods still make strong assumptions on the solutions that can be achieved. For instance \citet{yangqing11nips} optimizes binary pooling strategies that are given by the superposition of rectangular basis functions, and \citet{feng11cvpr} finds pooling regions by applying a linear discriminant  analysis for individual pooling strategies and training a classifier afterwards.
Also as opposed to \citet{ranzato2010modeling}, we aim for discriminative pooling over large neighborhoods in the SPM fashion where the information about the image class membership is available during training.

\paragraph{Outline}
We question restrictions imposed by the above methods and suggest to learn pooling strategies under weaker assumptions. Indeed, our method discovers new pooling shapes that were not found previously as they were suppressed by the more restrictive settings. 

The generality that we are aiming for comes at the price of a high dimensional  parameters space. This manifests in a complex optimization problem that is more demanding on memory requirements as well as computations needs, not to mention a possibility of over-fitting. Therefore, we also discuss two approximations to our method. First approximation introduces a pre-pooling step and therefore reduces the spatial dimension of the codes. The second approximation divides the codes into a set of smaller batches (subset of codes) that can be optimized independently and therefore in parallel. 

Finally, we evaluate our method on the CIFAR-10 and show strong improvements over hand-crafted pooling schemes in the regime of small dictionaries where our more flexible model shows its capability to make best use of the representation by exploring spatial pooling strategies specific to each coordinate of the code. Despite the diminishing return, the performance improvements persist up to largest codes we have investigated. We also show strong classification performance on the CIFAR-100 dataset where our method outperforms, to the best of our knowledge, the state-of-the-art. 

\section{Method}
\label{section:method}
As opposed to the methods that use fixed spatial pooling regions in the object classification task \citep{lazebnik2006beyond, yang2009linear} our method  jointly optimizes both the classifier and the pooling regions. 
In this way, the learning signal available in the classifier can help shaping the pooling regions in order to arrive at better pooled features. 

\subsection{Parameterized pooling operator}

The simplest form of the spatial pooling is computing histogram over the whole image. This can be expressed as $\Sigma(\bs{U}) := \sum_{j=1}^M \bs{u}_j$, where $\bs{u}_j \in \mathbb{R}^K$ is a code (out of $M$ such codes) and an index $j$ refers to the spatial location that the code originates from\footnote{That is $j = (x,y)$ where $x$ and $y$ refer to the spatial location of the center of the extracted patch.}. A code is an encoded patch extracted from the image. 
The proposed method is agnostic to the patch extraction method and encoding scheme.
Since the pooling approach looses spatial information of the codes, 
\citet{lazebnik2006beyond} proposed to first divide the image into subregions, and afterwards to create pooled features by concatenating histograms computed over each subregion. There are two problems with such an approach: first, the division is largely arbitrary and in particular independent of the data; second, discretization artifacts occur as spatially nearby codes can belong to two different regions as the 'hard' division is made.

In this paper we address both problems by using a parameterized version of the pooling operator
\begin{equation}
\label{eq:parameterized_pooling_operator}
	\Theta_{\bs{w}}(\bs{U}) := \sum_{j=1}^M \bs{w}_j \circ \bs{u}_j
\end{equation}
 where $\bs{a} \circ \bs{b}$ is the element-wise multiplication. 
Standard spatial division of the image can be recovered from Formula \ref{eq:parameterized_pooling_operator} by setting the vectors $\bs{w}_j$ either to a vector of zeros $\bs{0}$, or ones $\bs{1}$. For instance, features obtained from dividing the image into 2 subregions can be recovered from $\Theta$ by concatenating two vectors:
$\sum_{j=1}^{\frac{M}{2}} \bs{1} \circ \bs{u}_j 
+ \sum_{j = \frac{M}{2} + 1}^M \bs{0} \circ \bs{u}_j$, and 
$\sum_{j=1}^{\frac{M}{2}} \bs{0} \circ \bs{u}_j  + \sum_{j = \frac{M}{2} + 1}^M \bs{1} \circ \bs{u}_j$, where $\left\{1, ..., \frac{M}{2}\right\}$ and $\left\{\frac{M}{2}+1, ..., M\right\}$ refer to the first and second half of the image respectively.

In general, let $\mathfrak{F} := \left\{ \Theta_{\bs{w}} \right\}_{\bs{w}}$ be a family of the pooling functions given by Eq. \ref{eq:parameterized_pooling_operator}, parameterized by the vector $\bs{w}$, and let $\bestTwo{\bs{w}}{l}$ be the 'best' parameter chosen from the family $\mathfrak{F}$ based on the initial configuration $l$ and a given set of images.\footnote{
We will show the learning procedure that can select such parameter vectors in the following subsection.}
First row of Figure \ref{fig:pooling_regions_viz} shows four initial configurations that mimic the standard 2-by-2 spatial image division. Every initial configuration can lead to different $\bestTwo{\bs{w}}{l}$ as it is shown in Figure \ref{fig:pooling_regions_viz}.
Clearly, the family $\mathfrak{F}$ contains all possible 'soft' and 'hard' spatial divisions of the image, and therefore can be considered as their generalization.

\subsection{Learnable pooling regions}
\label{subsection:learnable_pooling_regions}
In SPM architectures the pooling weights $\bs{w}$ are designed by hand, here we aim for
joint learning $\bs{w}$ together with the parameters of the classifier. Intuitively, the classifier during training has access to the classes that the images belong to, and therefore can shape the pooling regions. On the other hand, the method aggregates statistics of the codes over such learnt regions and pass them to the classifier allowing  to achieve higher accuracy. Such joint training of the classifier and the pooling regions can be done by adapting the backpropagation algorithm \citep{bishop, lecun1998efficient}, and so can be interpreted as a densely connected multilayer perceptron \citep{collobert2004links, bishop}.

Consider a sampling scheme and an encoding method producing $M$ codes each $K$ dimensional. Every coordinate of the code is an input layer for the multilayer perceptron. Then we connect every $j$-th input unit at the layer $k$ to the $l$-th pooling unit $a^k_l$ via the relation $w_{lj}^k u_j^k$. Since the receptive field of the pooling unit $a^k_l$ consists of all codes at the layer $k$, we have 
$a^k_l := \sum_{j=1}^M w_{lj}^k u_j^k$, and so in the vector notation 
\begin{equation}
\label{eq:pooling_neurons}
	\bs{a}_l := \sum_{j=1}^M \bs{w}_j^l \circ \bs{u}_j 
	= \Theta_{\bs{w}^l}(\bs{U})
\end{equation}
Next, we connect all pooling units with the classifier allowing the information to circulate between the pooling layers and the classifier.

Although our method is independent of the choice of a dictionary and an encoding scheme, in this work we use  K-means with triangle coding $f_k(\bs{x}) := \max\left\{ 0, \mu(\bs{z}) - z_k \right\}$ \citep{coates2010analysis}. 

Similarly, every multi-class classifier that can be interpreted in terms of an artificial neural network can be used. In our work we employ logistic regression. This classifier is connected to the pooling units via the formula
\begin{equation}
\label{eq:logistic_regression}
	J(\bs{\Theta}) := -\frac{1}{D} \sum_{i=1}^D\sum_{j=1}^C 
	\bs{1}\{y^{(i)} = j\} \log p(y^{(i)} = j | \bs{a}^{(i)}; \bs{\Theta})
\end{equation}
where $D$ denotes the number of all images, $C$ is the number of all classes, $y^{(i)}$ is a label assigned to the $i$-th input image, and $\bs{a}^{(i)}$ are responses from the 'stacked'  pooling units $[\bs{a}_l]_l$ for the $i$-th image\footnote{Providing the codes $\bs{U}^{(i)}$ are collected from the $i$-th image and $\bs{a}_l^{(i)} := \Theta_{\bs{w}^l}(\bs{U}^{(i)})$ then $\bs{a}^{(i)} := [\bs{a}^{(i)}_l]_l$.}. We use the logistic function to represent the probabilities: 
$	p(y = j | \bs{x}; \bs{\Theta}) := 
\frac{\exp(\bs{\theta}_j^T \bs{x})}{\sum_{l=1}^C \exp(\bs{\theta}_l^T\bs{x})}$.
Since the classifier is connected to the pooling units, our task is to learn jointly the pooling parameters $\bs{W}$ together with the classifier parameters $\bs{\Theta}$, where $\bs{W}$ is the matrix containing all pooling weights.

Finally, we use standard gradient descent algorithm that updates the parameters using the following fixed point iteration
\begin{equation}
	\bs{X}^{t+1} := \bs{X}^t - \gamma \nabla J(\bs{X}^t)
\end{equation} 
where in our case $\bs{X}$ is a vector consisting of the pooling parameters $\bs{W}$ and the classifier parameters $\bs{\Theta}$. In practice, however, we employ a quasi-Newton algorithm LBFGS\footnote{The algorithm, developed by Mark Schmidt, can be downloaded from the following webpage: http://www.di.ens.fr/~mschmidt/Software/minFunc.html}. 

\subsection{Regularization terms}
\label{subsection:regularization}
In order to improve the generalization, we introduce regularization of our network as we deal with a large number of the parameters. 
For the classification $\bs{\Theta}$  and pooling parameters $\bs{W}$, we employ a simple $L_2$ regularization  terms: $\twoNorm{\bs{\Theta}}^2$ and $\sum_k\twoNorm{\bs{W}^k}^2$.
We improve the interpretability of the pooling weights as well as to facilitate a transfer among models by adding a projection onto a unit cube. 
To reduce quantization artifacts of the pooling strategy as well as to ensure smoothness of the output w.r.t. small translations of the image, the model penalizes weights whenever the pooling surface is non-smooth. This can be done by measuring the spatial variation, that is $\twoNorm{\nabla_x \bs{W}^k}^2 + \twoNorm{\nabla_y \bs{W}^k}^2$ for every layer $k$. This regularization enforces soft transition between the pooling subregions.

Every regularization term comes with its own hyper-parameter set by cross-validation. The overall objective that we want to optimize is 
\begin{align}
\label{eq:learnable_pooling_regions_objective}
&	\minimize_{\bs{W}, \bs{\Theta}}  J_{\text{R}}(\bs{\Theta}, \bs{W}) := \\ \nonumber
&		-\frac{1}{D} \sum_{i=1}^D\sum_{j=1}^C \bs{1}\{y^{(i)} = j\} 
			\log p(y^{(i)} = j | \bs{a}^{(i)}; \bs{\Theta}) \\ \nonumber
&		+ \frac{\alpha_1}{2}\twoNorm{\bs{\Theta}}^2 
	  + \frac{\alpha_2}{2} \twoNorm{\bs{W}}^2 \\ \nonumber
&	 	+ \frac{\alpha_3}{2}\left(\twoNorm{\nabla_x \bs{W}}^2 
		+ 	\twoNorm{\nabla_y \bs{W}}^2\right) \\ \nonumber
		\\ \nonumber
&	\text{subject to} \; \bs{W} \in \left[0,1\right]^{K\times M\times L} \nonumber
\end{align}
where 
$\bs{a}_l$ is the $l$-th pooling unit described by Formula \ref{eq:pooling_neurons}, and $\twoNorm{\bs{W}}$ 
is the Frobenius norm.

\subsection{Approximation of the model}
\label{subsection:approximation_of_the_model}
The presented approach is demanding to train in the means of the CPU time and memory storage when 
using high dimensional representations. That is, the number of the pooling parameters to learn grows as $K\times M\times L$, where $K$ is dimensionality of codes, $M$ is the number of patches taken from the image and $L$ is the number of pooling units. Therefore, we propose two approximations to our method
making the whole approach more scalable towards bigger dictionaries. However, we emphasize that learnt pooling regions have very little if any overhead compared to standard spatial division approaches at test time.

First approximation does a fine-grained spatial partition of the image, and then pools the codes over such subregions. This operation, we call it a pre-pooling step, reduces the number of considered spatial locations by the factor of the pre-pooling size. For instance, if we collect $M$ codes and the pre-pooling size is $S$ per dimension, then we reduce the number of codes to a number $\frac{M}{S^2}$. The pre-pooling operation fits well into our generalization of the SPM architectures as by choosing $S := \frac{M}{2}$ we obtain a weighted quadrants scheme. Moreover, the modeler has the option to start with the larger $S$ when little data is available and gradually decreases $S$ as more parameters can be learnt using more data.

The second approximation divides a $K$ dimensional code into $\frac{K}{D}$ batches, each $D$ dimensional (where $D \leq K$ and $K$ is divisible by $D$). Then we train our model on all such batches in parallel to obtain the pooling weights. Later, we train the classifier on top of the concatenation of the trained, partial models.
As opposed to \citet{le2011building} our training is fully independent and doesn't need communication between different machines.

Since the ordering of the codes is arbitrary, we also consider $D$ dimensional batches formed from the permuted version of the original codes, and combine them together with the concatenated batches to boost the classification accuracy (we call this approximation redundant batches). Given a fixed sized dictionary, this approximation  performs slightly better, although it comes at the cost of increased number of features due to the redundant batches.

Finally, our approximations not only lead to a highly parallel training procedure with reduced memory requirements and computational demands, but also have shown to greatly reduce the number of required iterations as they tend to converge roughly $5$ times faster than the full model on large dictionaries. 

\section{Experimental Results}
\label{section:experimental_results}

We evaluate our method on the CIFAR-10 and CIFAR-100 datasets \citep{krizhevsky2009learning}. Furthermore, we provide insights into the learnt pooling strategies as well as investigate transfer between datasets.
In this section we describe our experimental setup, and present our results on both datasets.

\subsection{CIFAR-10 and CIFAR-100 datasets}
The CIFAR-10 and CIFAR-100 datasets contain $50000$ training color images and $10000$ test color images from respectively $10$ and $100$ categories, with $6000$ and $600$ images per class respectively. All images have the same size: $32\times32$ pixels, and were sampled from the 80 million tiny images dataset \citep{torralba200880}. 

\subsection{Evaluation pipeline}
In this work, we follow the \citet{coates2011importance} pipeline. 
We extract normalized
and whitened $6\times6$ patches from images using a dense, equispaced grid with a unit sample spacing. As the next step, we employ the K-means assignment and triangle encoding \citep{coates2011importance, coates2010analysis} to compute codes -- a K-dimensional representation of the patch.
We classify images using either a logistic regression, or a linear SVM in the case of transferred pooling regions. Optionally we use two approximations described in subsection \ref{subsection:approximation_of_the_model}.  As we want to be comparable to \citet{coates2010analysis}, who use a spatial division into 2-by-2 subregions which results in $4 \cdot K$ pooled features,
we use $4$ pooling units.
Furthermore, we use standard division (first row of Figure \ref{fig:pooling_regions_viz}) as an initialization of our model.

To learn parameters of the model we use the limited-memory BFGS algorithm (details are described in subsection \ref{subsection:learnable_pooling_regions}), and limit the number iterations to $3000$. After the training, we can also concatenate the results of the parameterized pooling operator  $\left[\Theta_{\bs{w}_l}(\bs{U})\right]_{l=1}^4$. This yields a $4 \cdot K$ dimensional feature vector that can be again fed into the classifier, and trained independently with the already trained pooling regions. We call this procedure transfer of pooling regions. 

The reason behind the transfer is threefold. Firstly, we can combine partial models trained with our approximation in batches to a full, originally intractable, model\footnote{The reader can find details of such approximation in subsection \ref{subsection:approximation_of_the_model}.}.
Secondly, the transfer process allows to combine both the codes and the learnt model from the dictionaries of different sizes.
Lastly, it enables training of the pooling regions together with the classifier on one dataset, and then re-train the classifier alone on a target dataset.
To transfer the pooling regions, we tried logistic regression classifier and linear SVM showing that both classifying procedures can benefit from the learnt pooling regions. However, since we achieve slightly better results for the linear SVM (about $0.5\%$ for bigger dictionaries), only those results are reported. Similarly, we don't notice significant difference in the classification accuracy for smaller dictionaries when the pre-pooling is used (with the pre-pooling size $S := 3$), and therefore all experiments refer only to this case.
Finally, we select hyper-parameters of our model based on the $5$-fold cross-validation.

\subsection{Evaluation of our method on small dictionaries}
\begin{figure}[t]
\begin{center}
\subfigure[]{
\label{fig:coates_against_pooling_regions}
\includegraphics[width=0.35\linewidth]{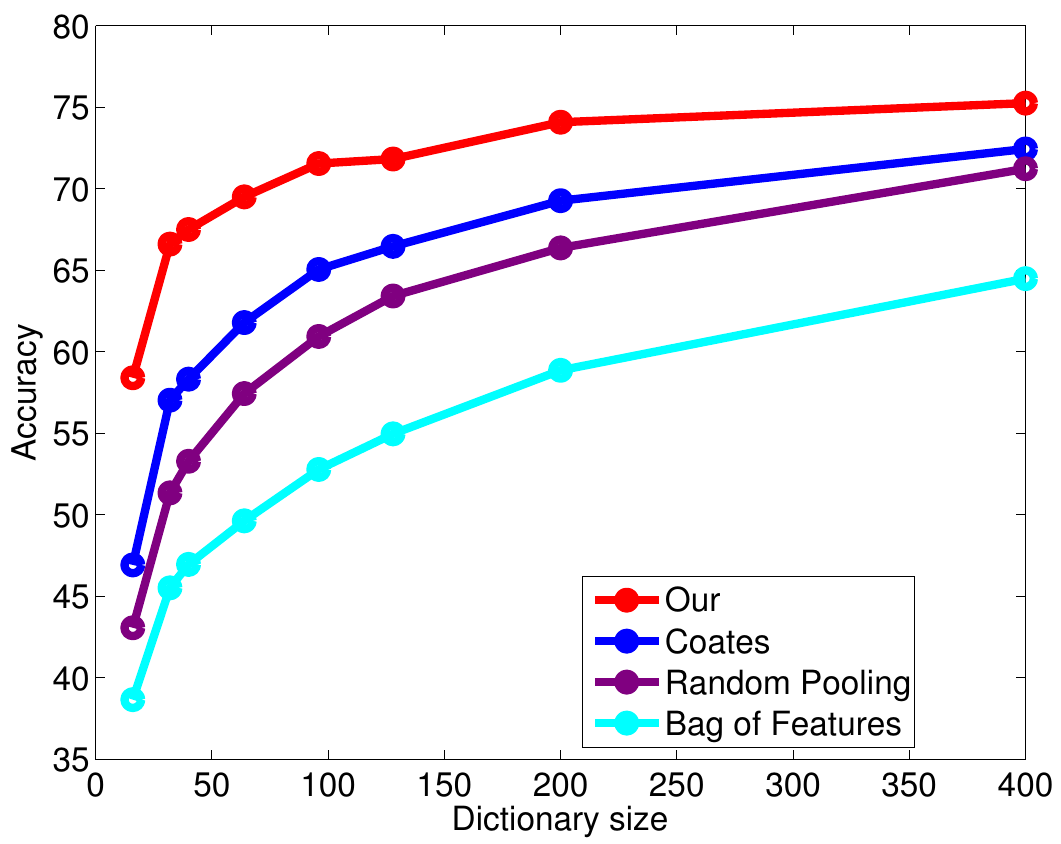}
}
\subfigure[]{
\label{fig:coates_against_batched_pooling_regions}
\includegraphics[width=0.35\linewidth]{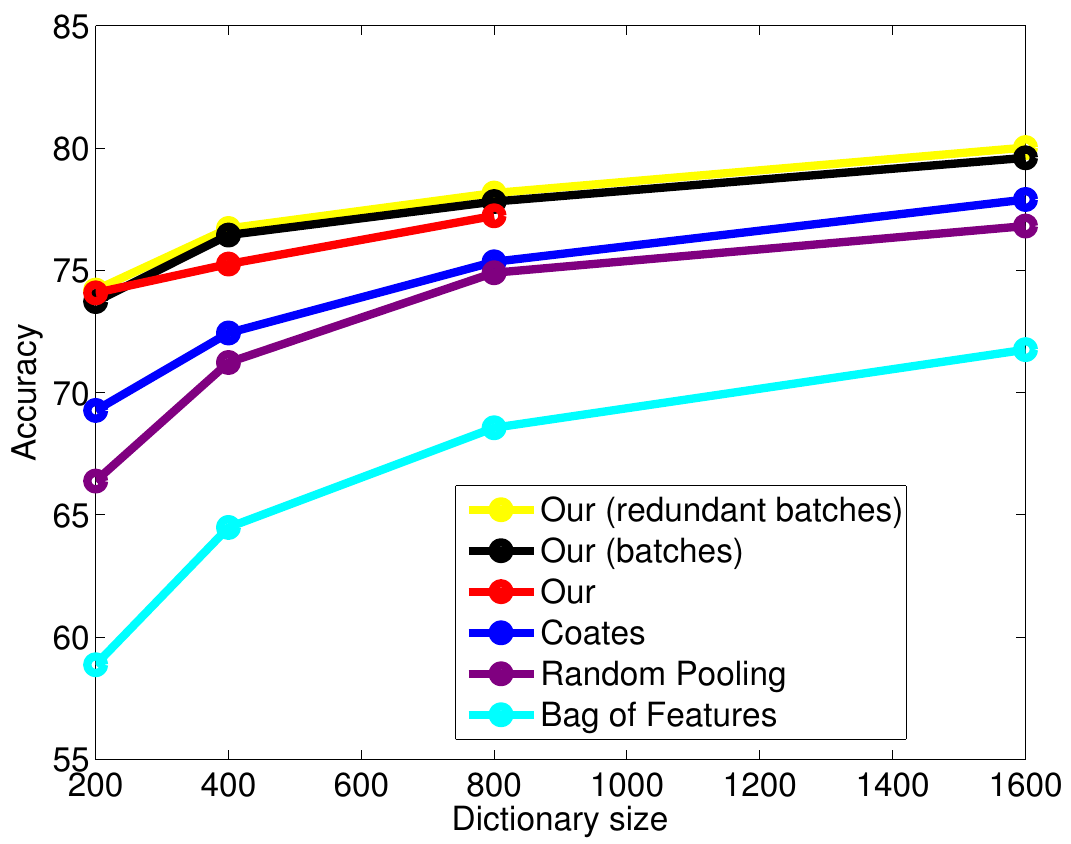}
}
\end{center}
\caption{Figure \ref{fig:coates_against_pooling_regions} shows accuracy of the classification with respect to the number of dictionary elements on smaller dictionaries. Figure 
\ref{fig:coates_against_batched_pooling_regions} shows the accuracy of the classification for bigger dictionaries when batches, and the redundant batches were used. Experiments are done on CIFAR-10.}
\label{fig:accuracy_figure}
\end{figure}

Figure \ref{fig:coates_against_pooling_regions} shows the classification accuracy of our full method against the baseline \citep{coates2011importance}. Since we train the pooling regions without any approximations in this set of experiments the results are limited to dictionary sizes up to 800. Our method outperforms the approach of Coates by $10\%$ for dictionary size $16$ (our method achieves the accuracy $57.07\%$, whereas the baseline only $46.93\%$). This improvement is consistent up to the bigger dictionaries although the margin is getting smaller. Our method is about $2.5\%$ and $1.88\%$ better than the baseline for $400$ and $800$ dictionary elements respectively. 

\subsection{Scaling up to sizable dictionaries}
In subsection \ref{subsection:approximation_of_the_model} we have discussed the possibility of dividing the codes into low dimensional batches  and learning the pooling regions on those. In the following experiments we use batches with $40$ coordinates extracted from the original code, as those fit conveniently into the memory of a single, standard machine (about $5$ Gbytes for the main data) and can all be trained in parallel. 

Besides a reduction in the memory requirements, the batches have shown multiple benefits in practice due to smaller number of parameters. We need less computations per iterations as well as observe faster convergence. Figure \ref{fig:coates_against_batched_pooling_regions} shows the classification performance for larger dictionaries where we examined the full model [Our], the baseline [Coates], random pooling regions (described in subsection \ref{subsection:random_pooling_regions}), bag of features, and two possible approximation - the batched model [Our (batches)], and the redundantly batched model [Our (redundant batches)].

Our test results are presented in Table \ref{tab:coates_pooling_experiments}. 
When comparing our full model to the approximated versions with batches for dictionaries of size 200, 400 and 800, we observe that there is almost no drop in performance and we even slightly improve for the bigger dictionaries. We attribute this to the better conditioned learning problem of the smaller codes within one batch.
With an accuracy for the batched model of $79.6\%$ we outperform the Coates baseline by $1.7\%$. Interestingly, we gain another small improvement to $80.02\%$ by adding redundant batches which amounts to a total improvement of $2.12\%$ compared to the baseline. Our method performs comparable to the pooling strategy of \citet{yangqing11nips} which uses more restrictive assumptions on the pooling regions and employs feature selection algorithm.

\begin{table}[H]
\centering
\begin{tabular}{| l | c | c | r |}
	\hline
	Method & Dict. size & Features & Acc. \\
	\hline 
		Jia & $1600$ & $6400$ & $80.17\%$ \\
\hline
 	Coates & $1600$ & $6400$ & $77.9\%$ \\
	Our (batches) & $1600$ & $6400$ & $79.6\%$\\
	Our (redundant) & $1600$ & $12800$ & $80.02\%$\\
	\hline
\end{tabular}
\caption{
Comparison of our methods against the baseline \citep{coates2011importance} and \citet{yangqing11nips} with respect to the dictionary size, number of features and the test accuracy on CIFAR-10.
}
\label{tab:coates_pooling_experiments}
\end{table}

To the best of our knowledge \citet{ciresan2012multi} achieves the best results on the CIFAR-10 dataset with an accuracy $88.79\%$ with a method based on a deep architecture -- different type of architecture to the one that we investigate in our study. More recently \citet{Goodfellow_maxout_2013} has achieved accuracy $90.62\%$ with new maxout model that takes an advantage of dropout.

\subsection{Random pooling regions}
\label{subsection:random_pooling_regions}
Our investigation also includes results using random pooling regions where the weights for the parameterized operator (Eq. \ref{eq:pooling_neurons}) were sampled from normal distribution with mean $0.5$ and standard deviation $0.1$, that is 
$\bs{w}_j^l \sim \mathcal{N}\left(0.5, 0.1\right)$ for all $l$.
This notion of the random pooling differs from the \citet{jia2012beyond} where random selection of rectangles is used.
The experiments show that the random pooling regions can compete with the standard spatial pooling (Figure \ref{fig:coates_against_pooling_regions} and \ref{fig:coates_against_batched_pooling_regions}) on the CIFAR-10 dataset, and suggest that random projection can still preserve some spatial information. This is especially visible in the regime of bigger dictionaries where the difference is only $1.09\%$. The obtained results indicate that  hand-crafted division of the image into subregions is questionable, and call for a learning-based approach. 

\begin{table*}
\centering
\begin{tabular}{|c|cccc|cccc|}\hline

regularization & \multicolumn{8}{|c|}{pooling weights}\\\hline

\multicolumn{1}{|c}{}  & \multicolumn{8}{c|}{dataset: CIFAR-10 ; dictionary size: 200}\\\hline

Coates (no learn.) &
\includegraphics[width=0.07\linewidth]{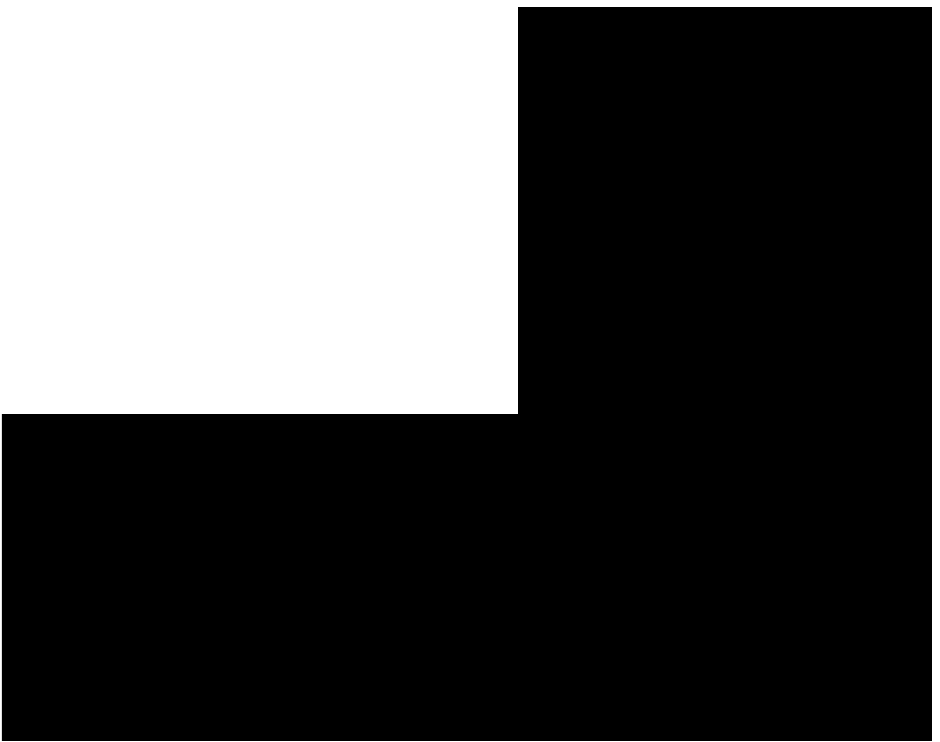} &
\includegraphics[width=0.07\linewidth]{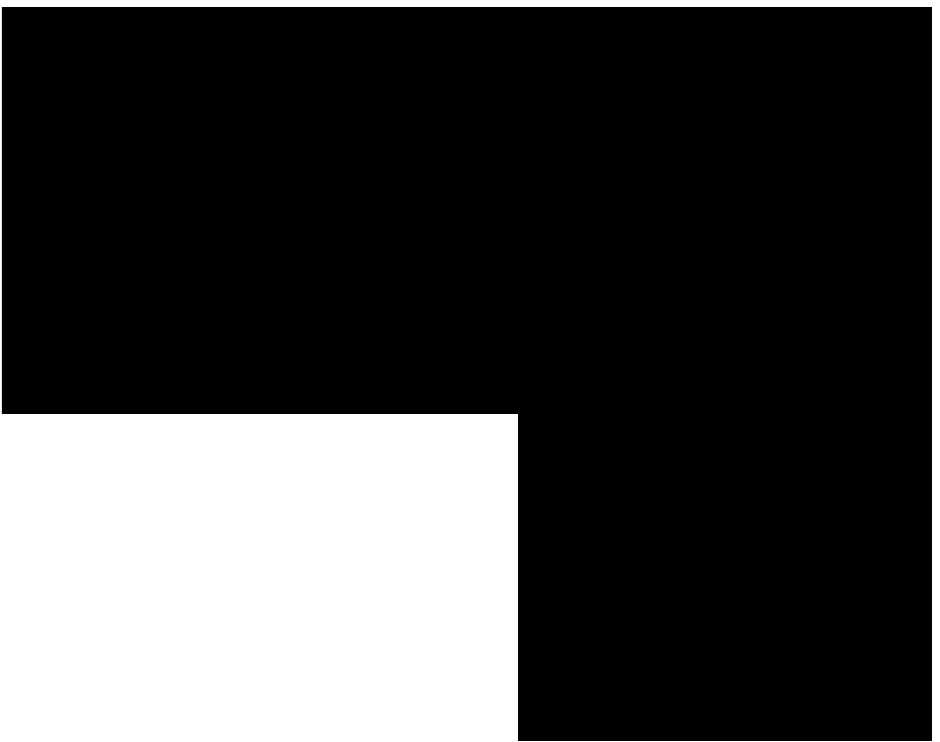} &
\includegraphics[width=0.07\linewidth]{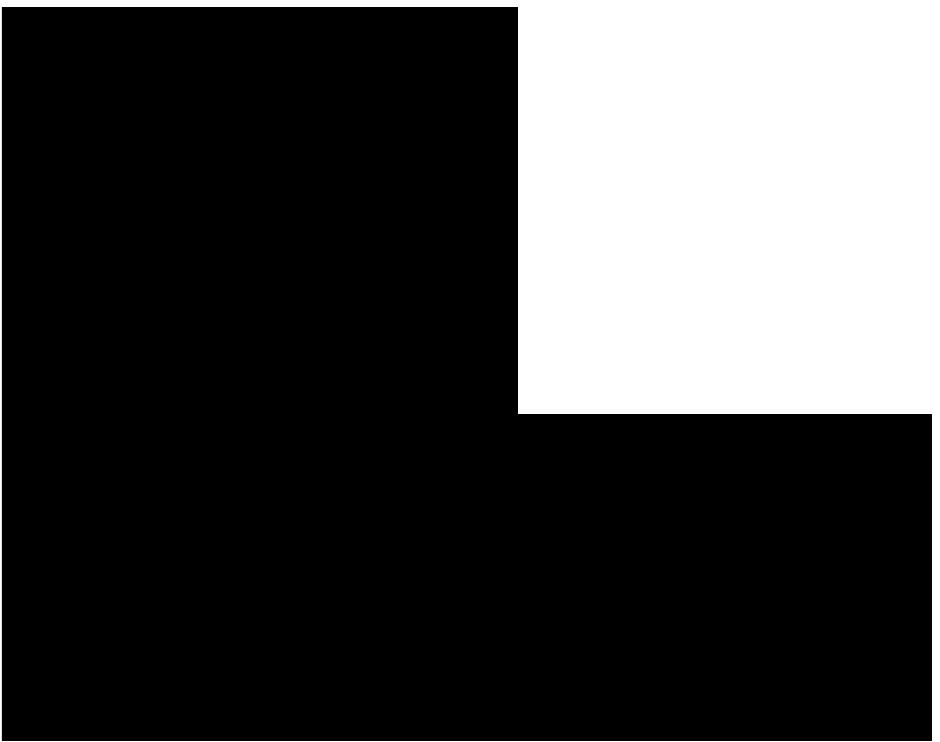} &
\includegraphics[width=0.07\linewidth]{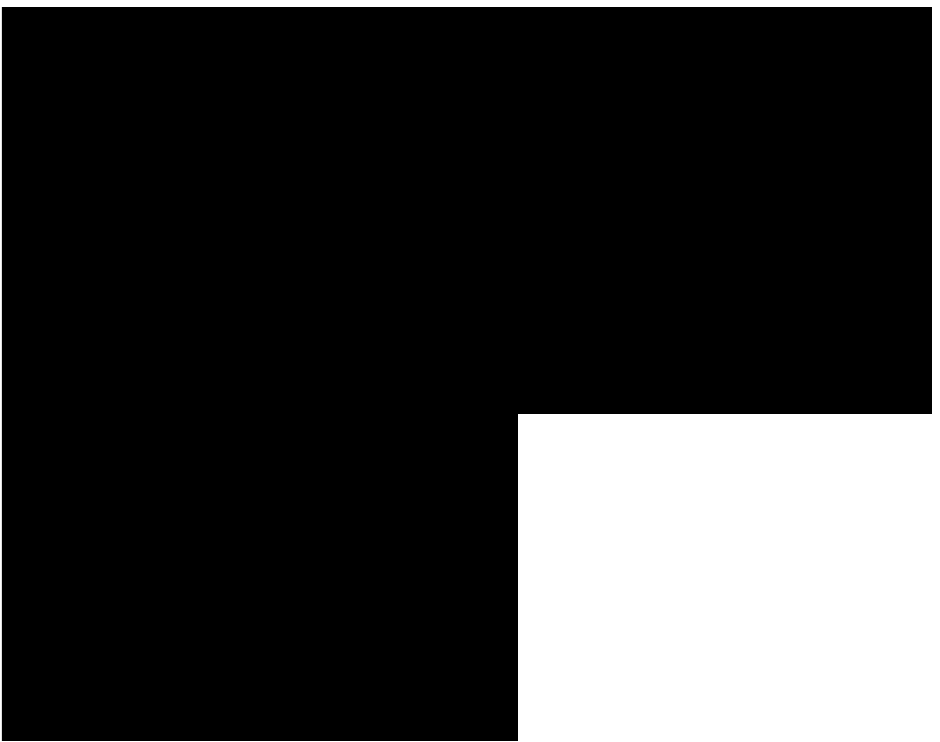} &
\includegraphics[width=0.07\linewidth]{images/weights/CIFAR-10/icml_submission/poolingRegionsInit_poolingNeuronNo1_layerNo10} &
\includegraphics[width=0.07\linewidth]{images/weights/CIFAR-10/icml_submission/poolingRegionsInit_poolingNeuronNo2_layerNo10} &
\includegraphics[width=0.07\linewidth]{images/weights/CIFAR-10/icml_submission/poolingRegionsInit_poolingNeuronNo3_layerNo10} &
\includegraphics[width=0.07\linewidth]{images/weights/CIFAR-10/icml_submission/poolingRegionsInit_poolingNeuronNo4_layerNo10} \\\hline

l2 & 
\includegraphics[width=0.07\linewidth]{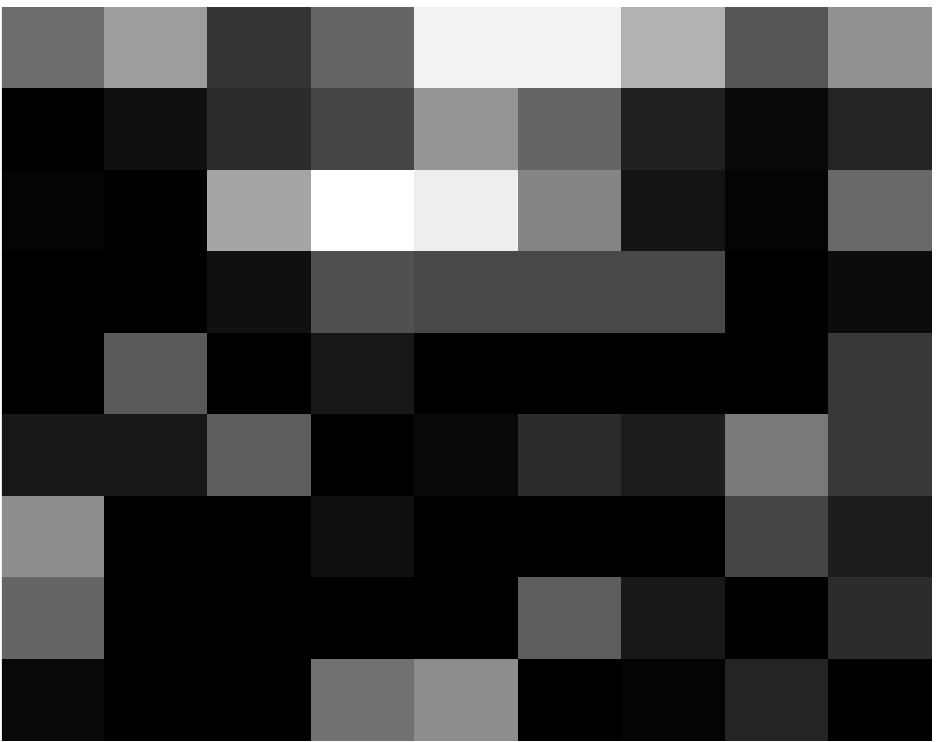} &
\includegraphics[width=0.07\linewidth]{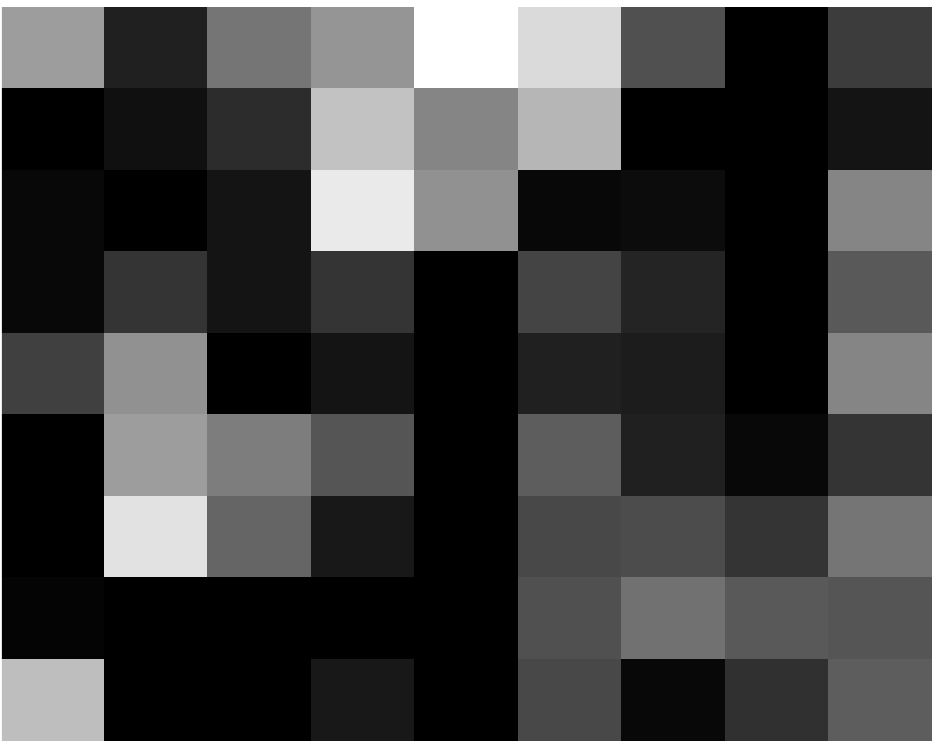} & 
\includegraphics[width=0.07\linewidth]{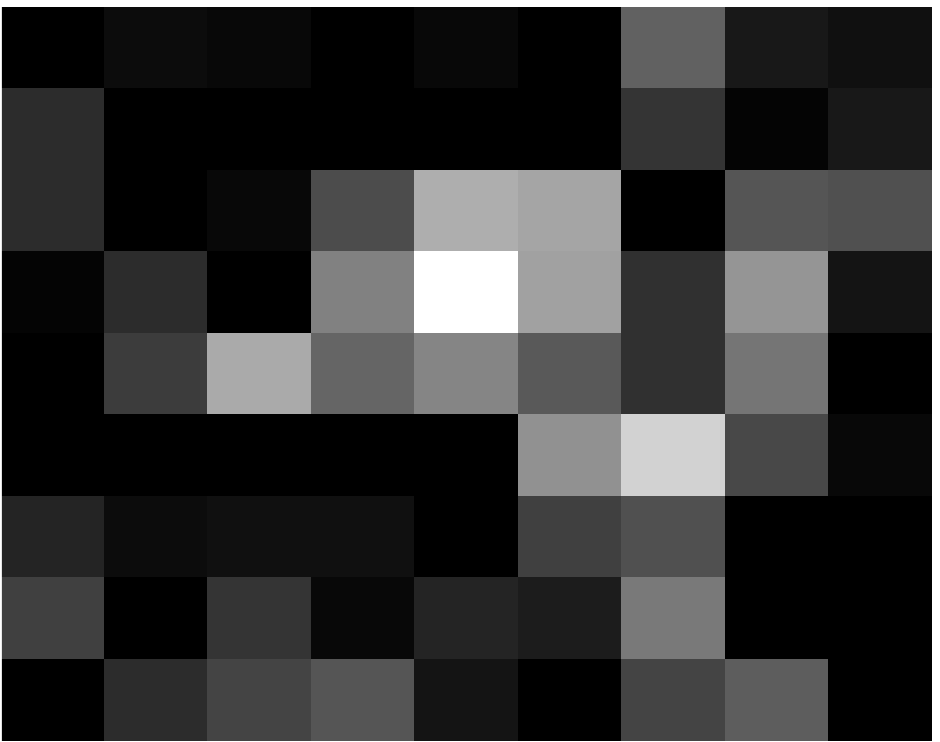} &
\includegraphics[width=0.07\linewidth]{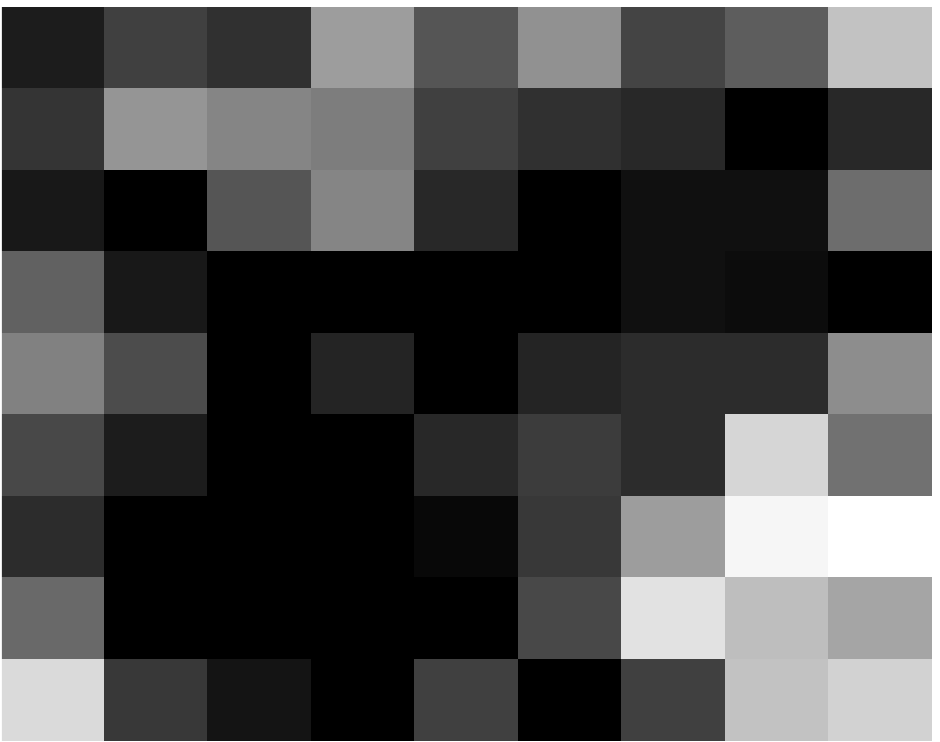} &
\includegraphics[width=0.07\linewidth]{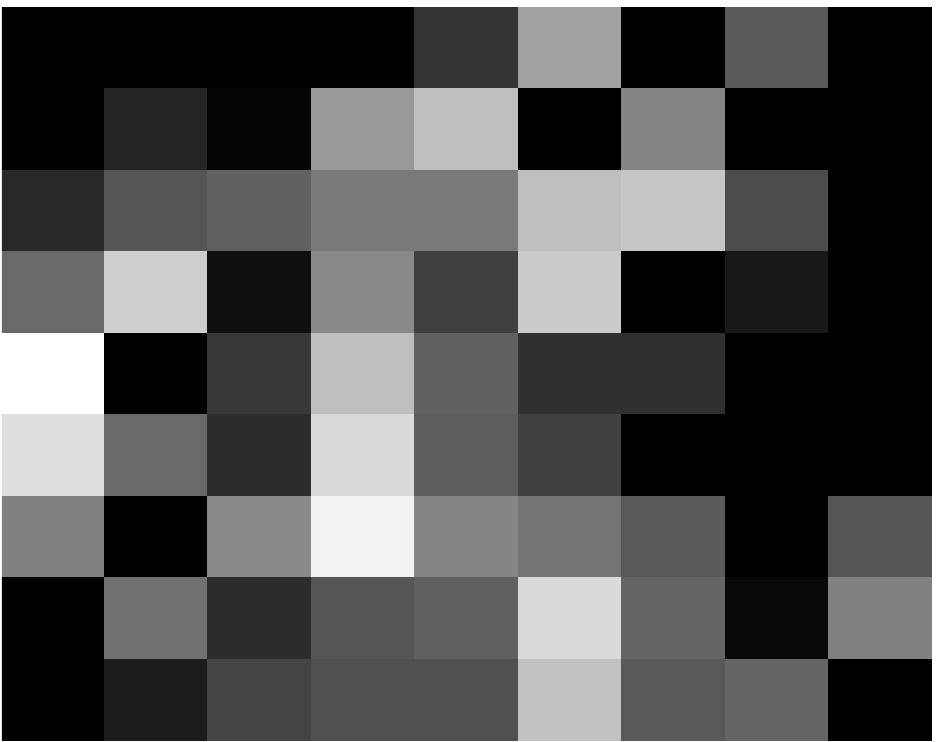} &
\includegraphics[width=0.07\linewidth]{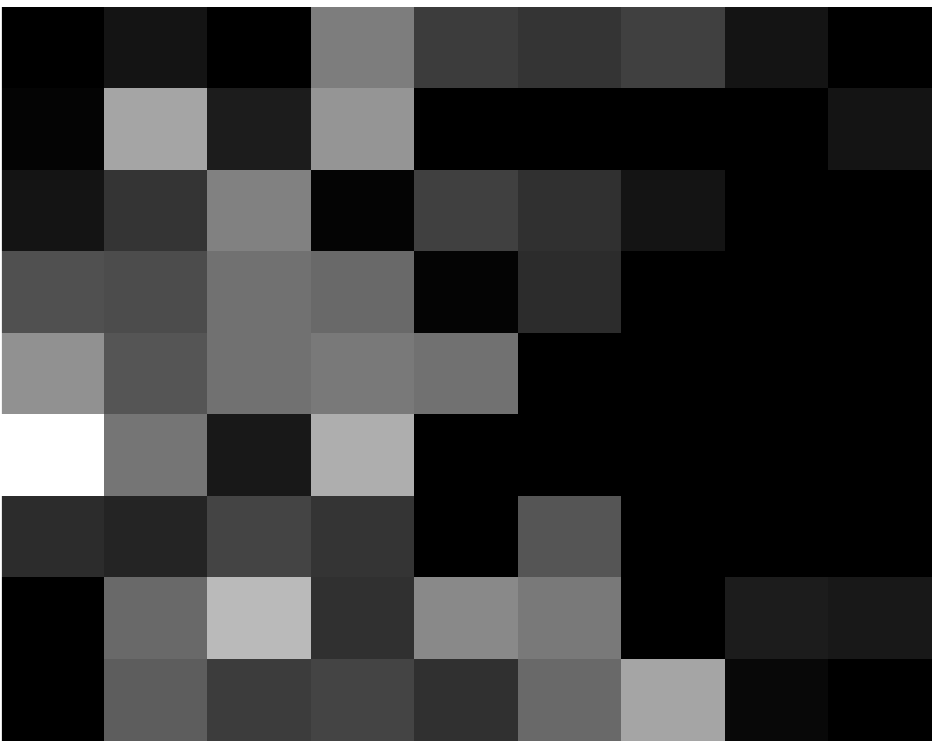} & 
\includegraphics[width=0.07\linewidth]{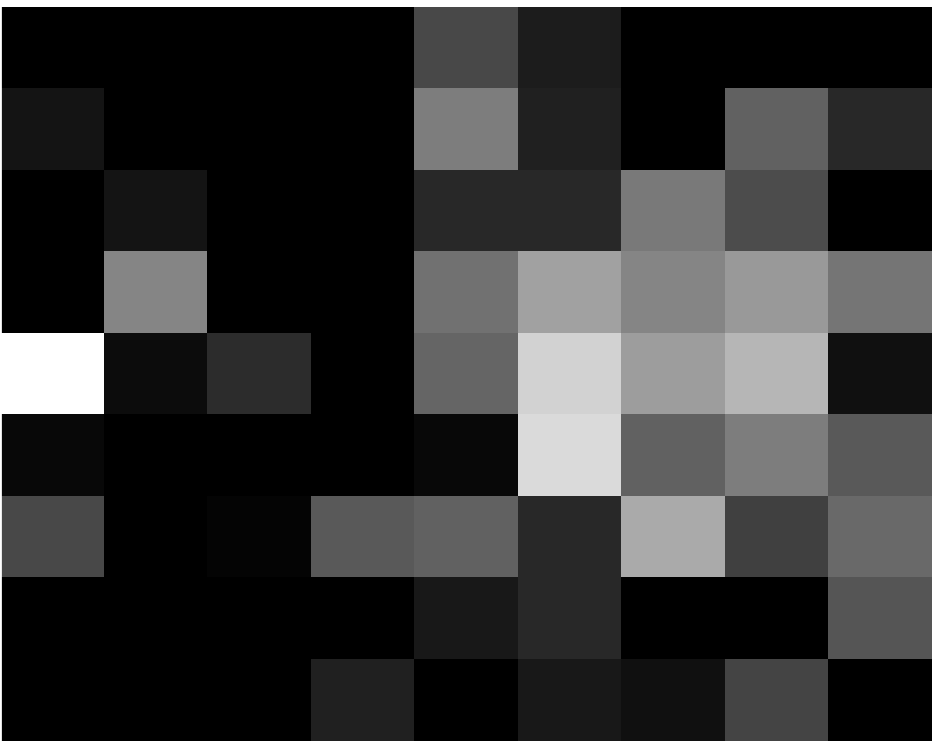} &
\includegraphics[width=0.07\linewidth]{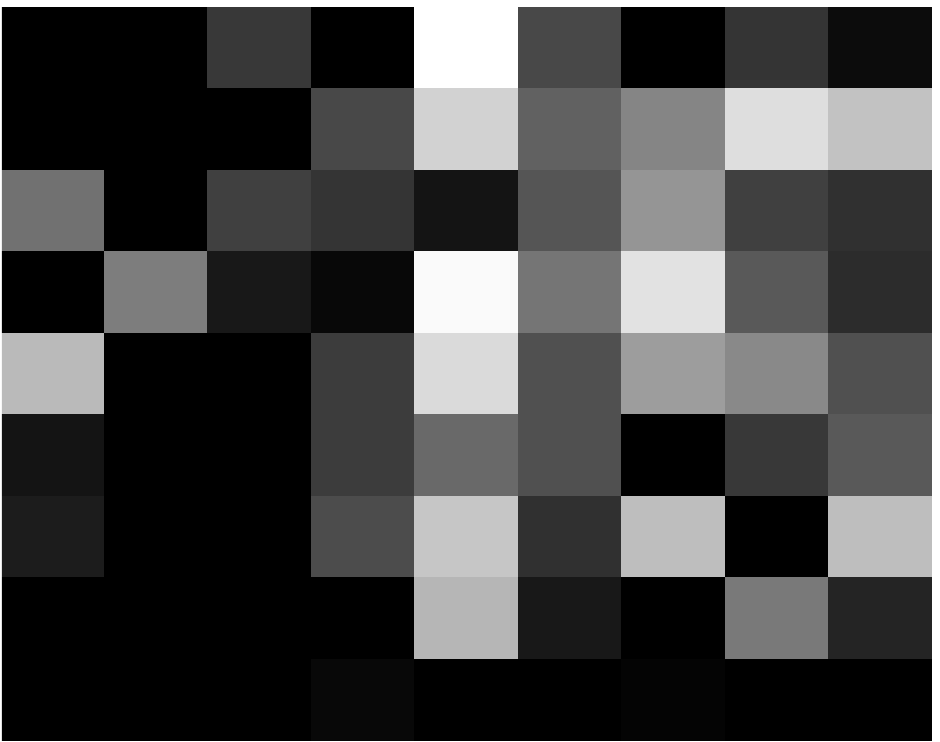} \\\hline

smooth & 
\includegraphics[width=0.07\linewidth]{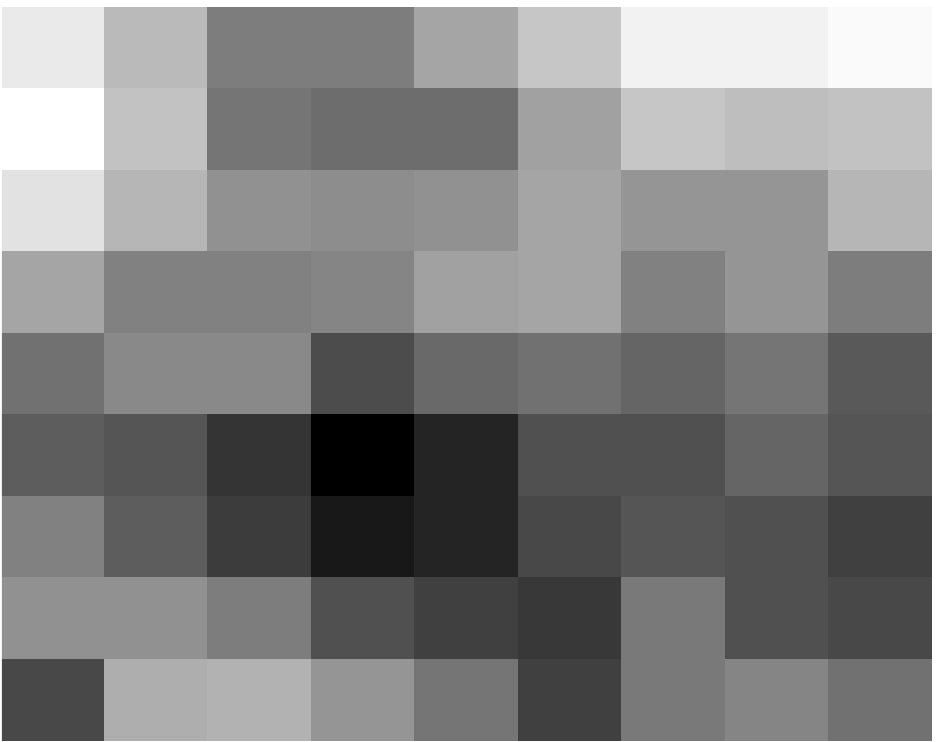} &
\includegraphics[width=0.07\linewidth]{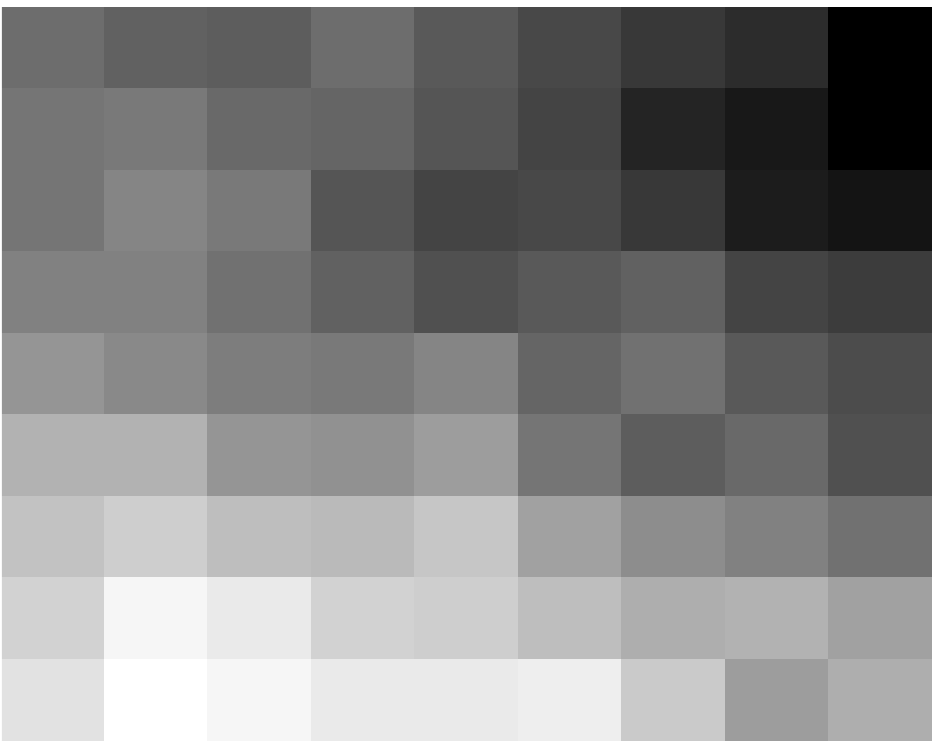} & 
\includegraphics[width=0.07\linewidth]{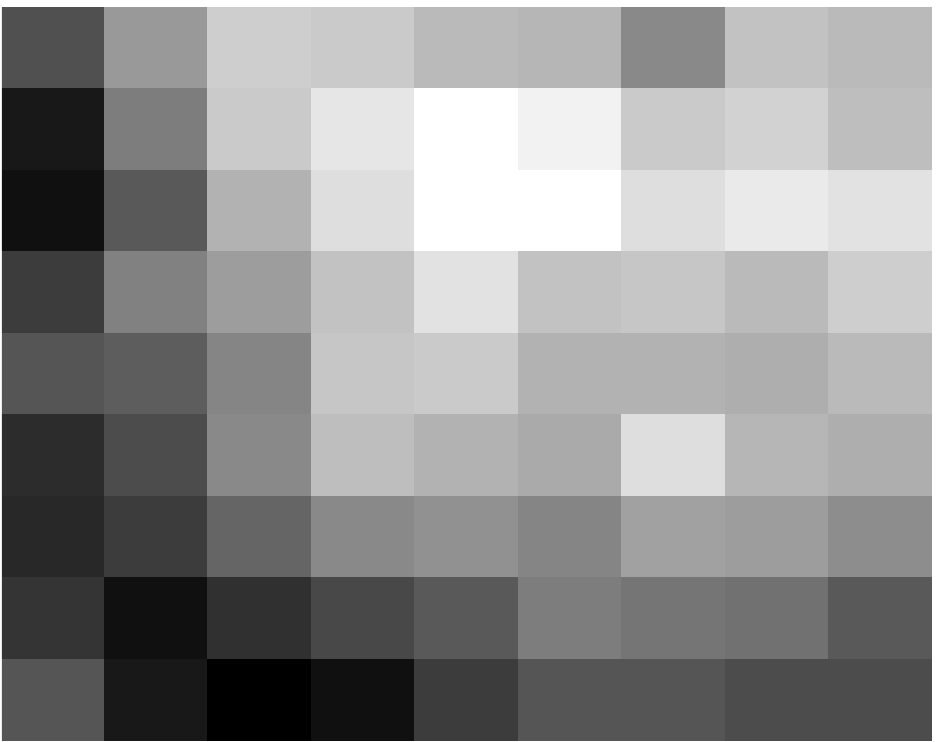} &
\includegraphics[width=0.07\linewidth]{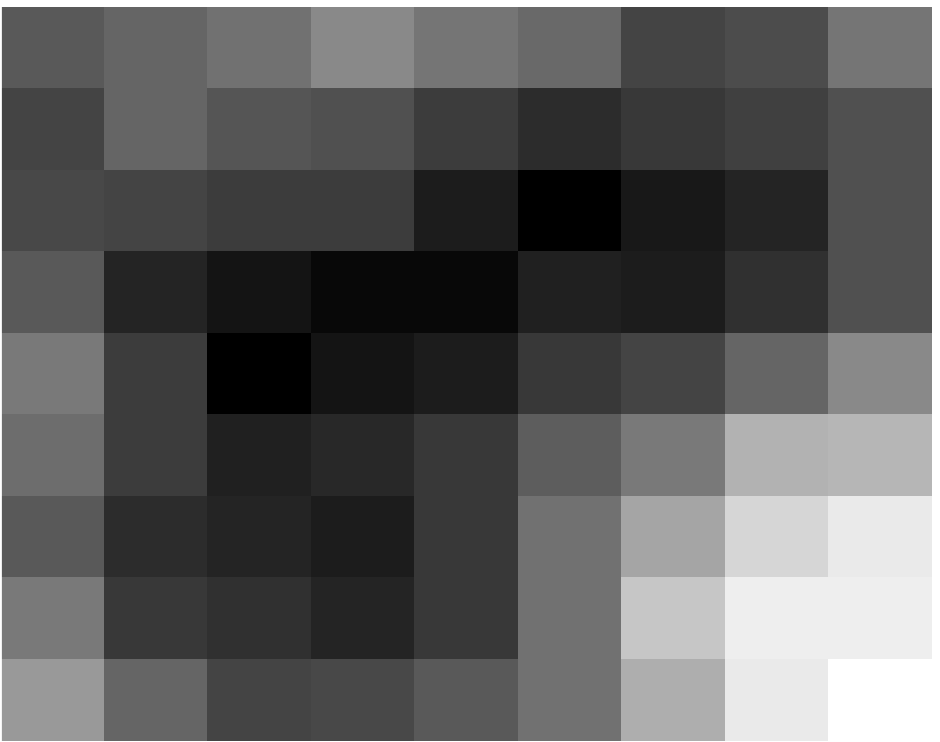} &
\includegraphics[width=0.07\linewidth]{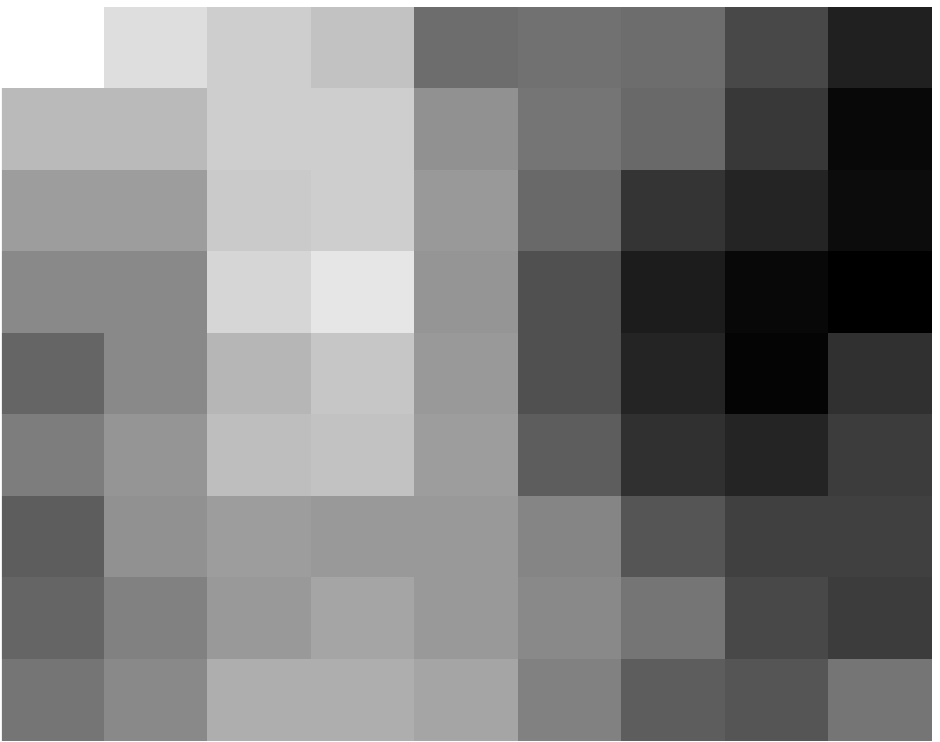} &
\includegraphics[width=0.07\linewidth]{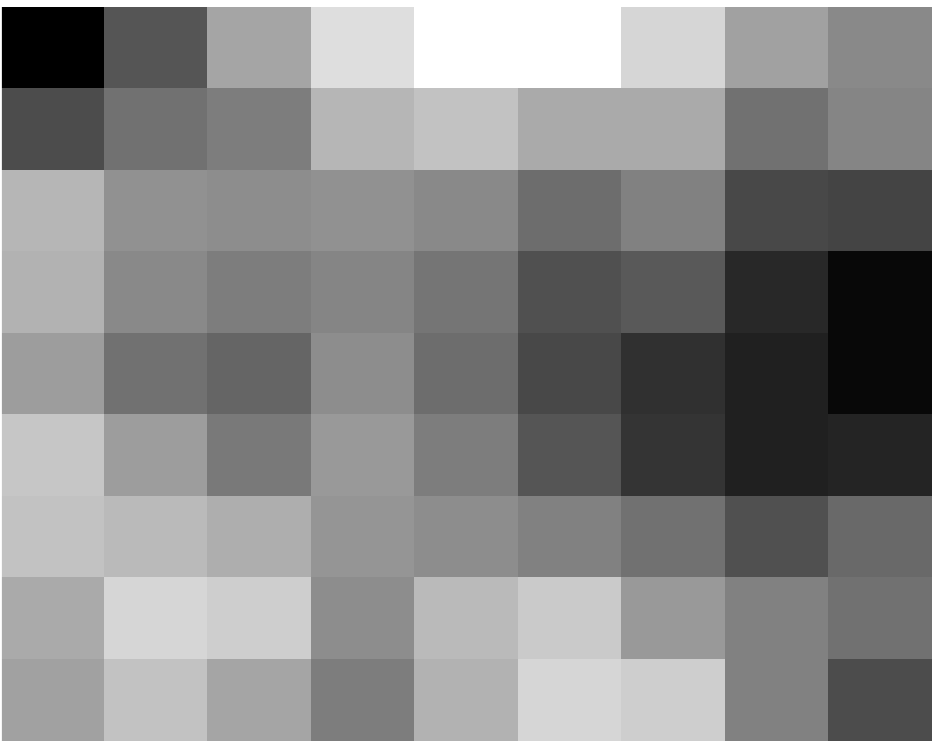} & 
\includegraphics[width=0.07\linewidth]{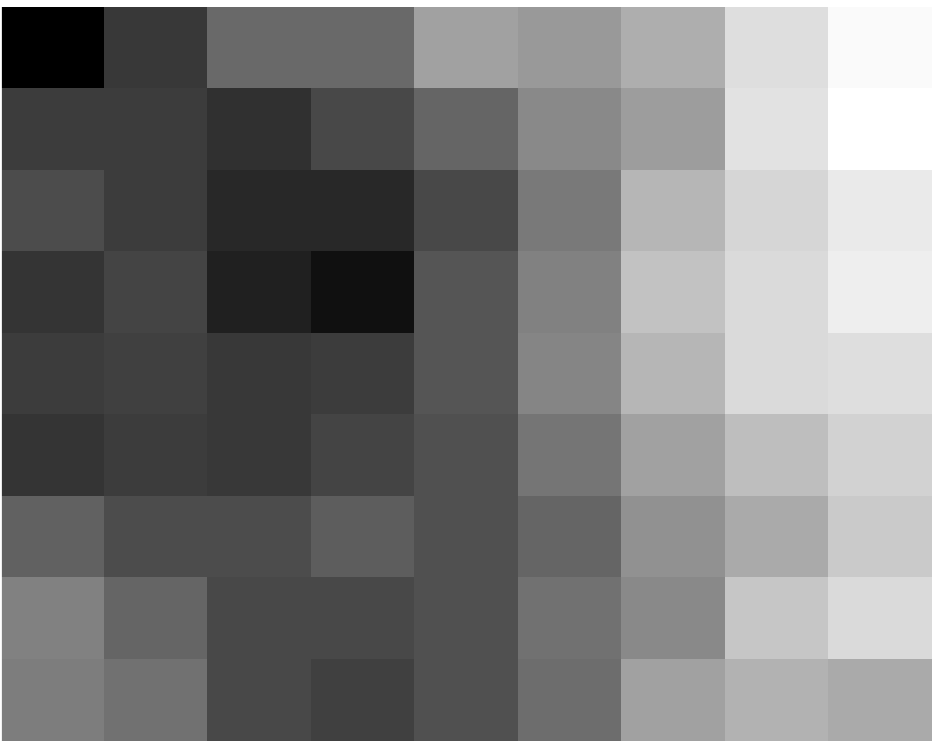} &
\includegraphics[width=0.07\linewidth]{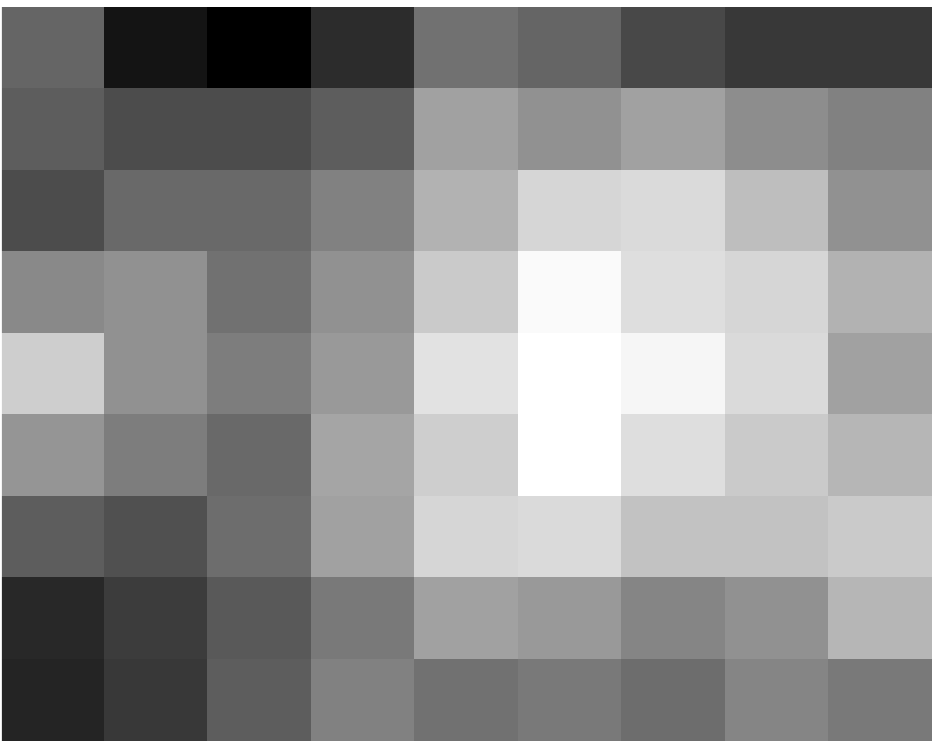} \\\hline

smooth \& l2 &
\includegraphics[width=0.07\linewidth]{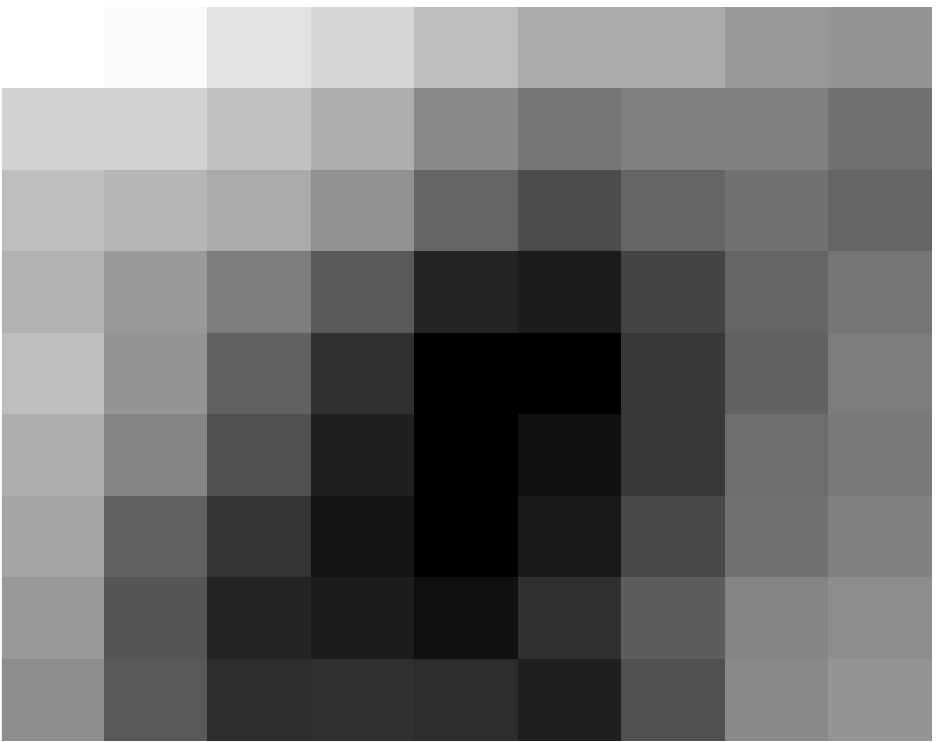} &
\includegraphics[width=0.07\linewidth]{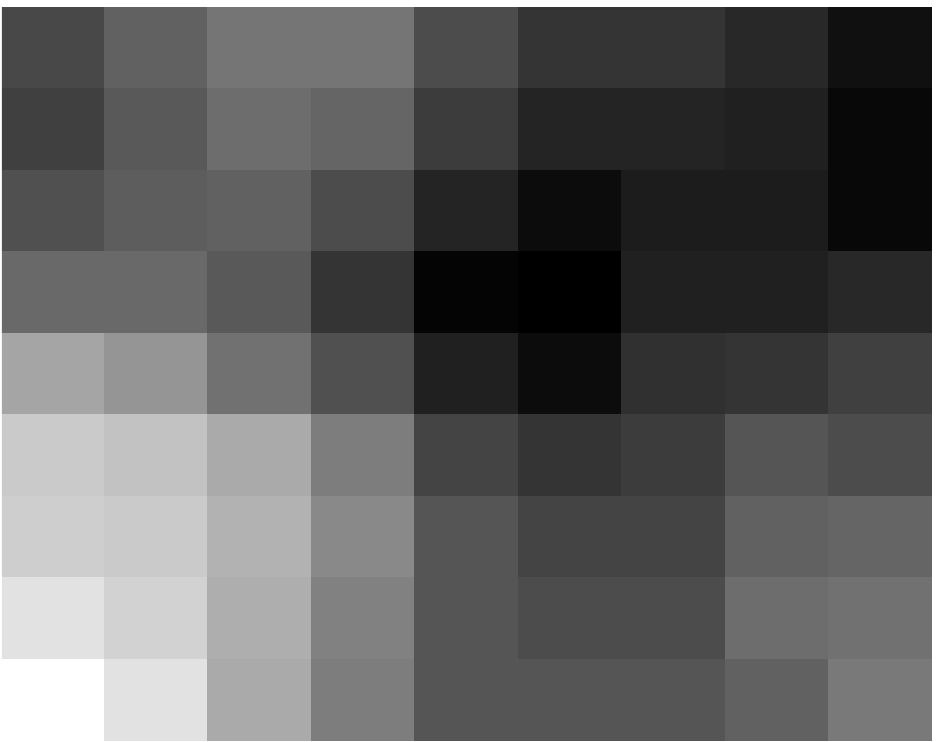} & 
\includegraphics[width=0.07\linewidth]{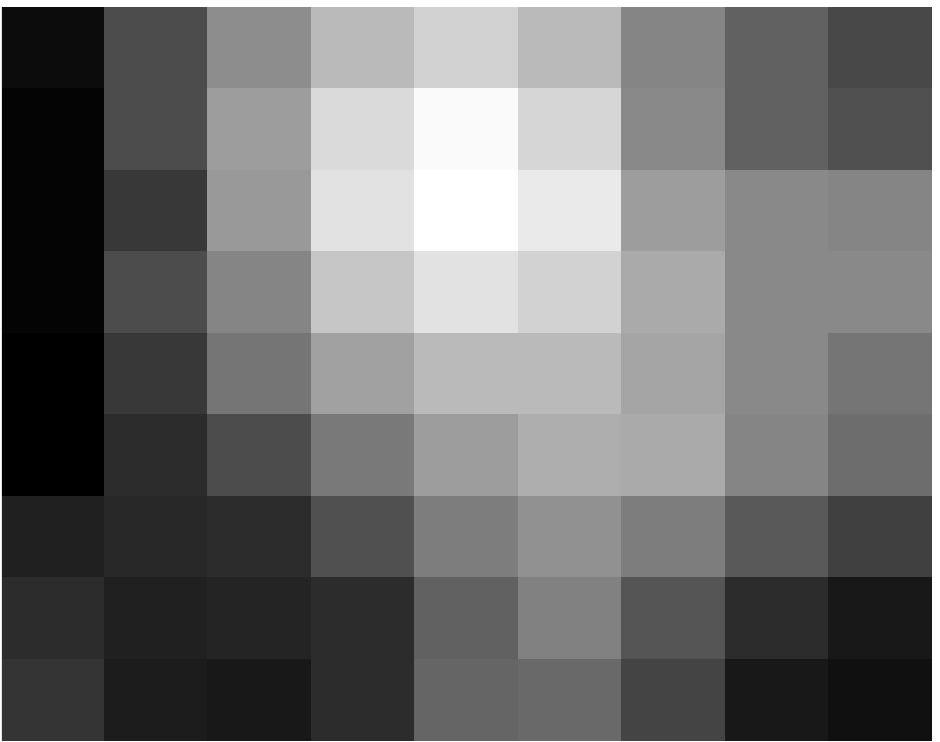} &
\includegraphics[width=0.07\linewidth]{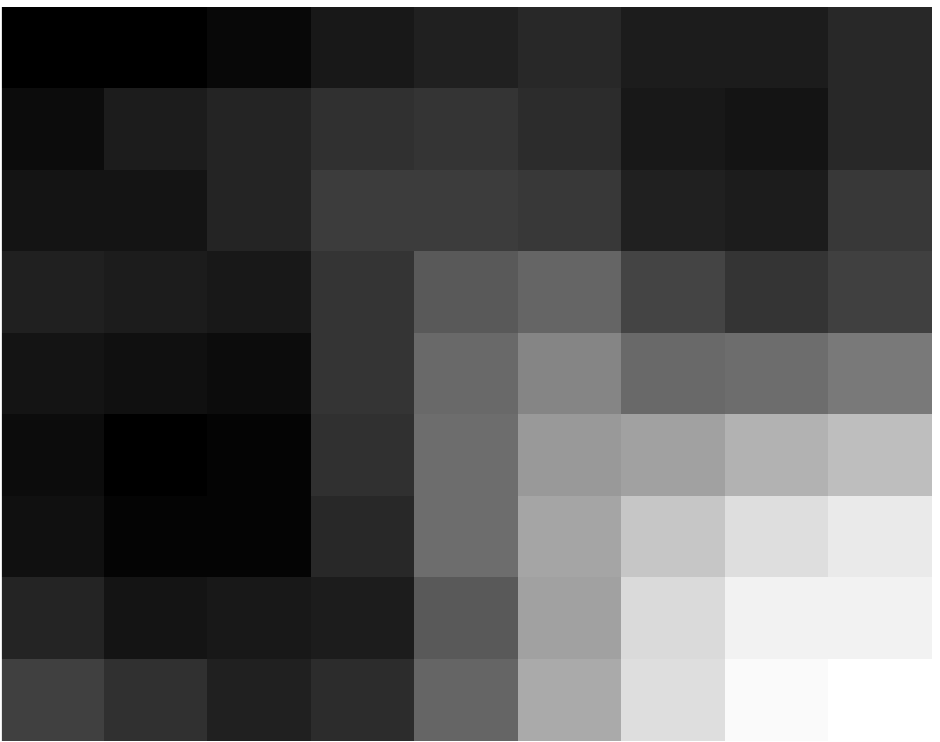} &
\includegraphics[width=0.07\linewidth]{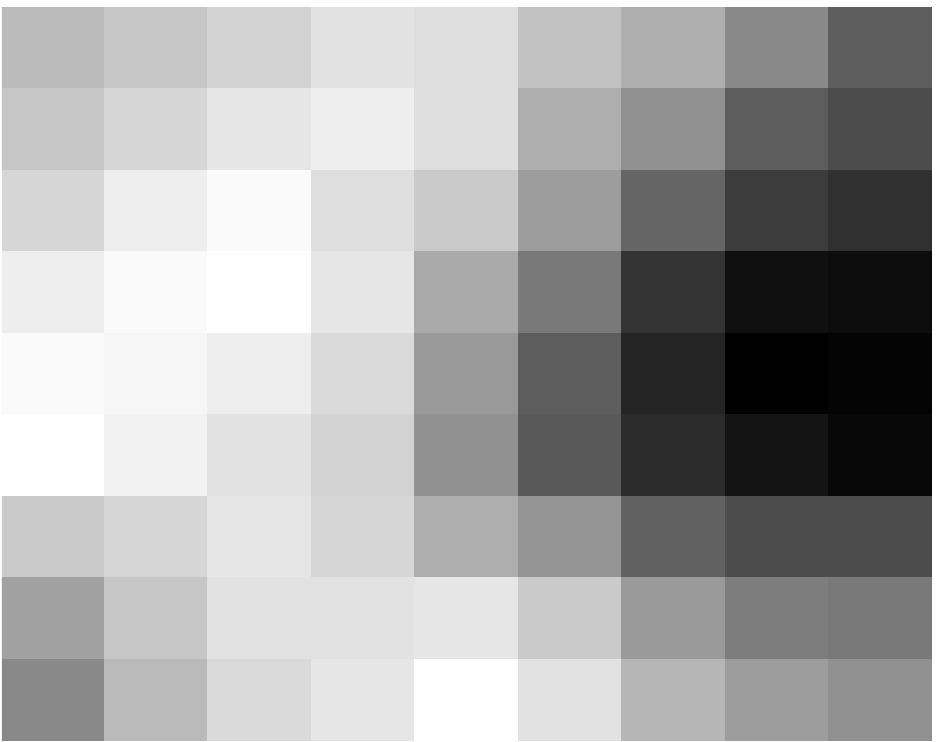} &
\includegraphics[width=0.07\linewidth]{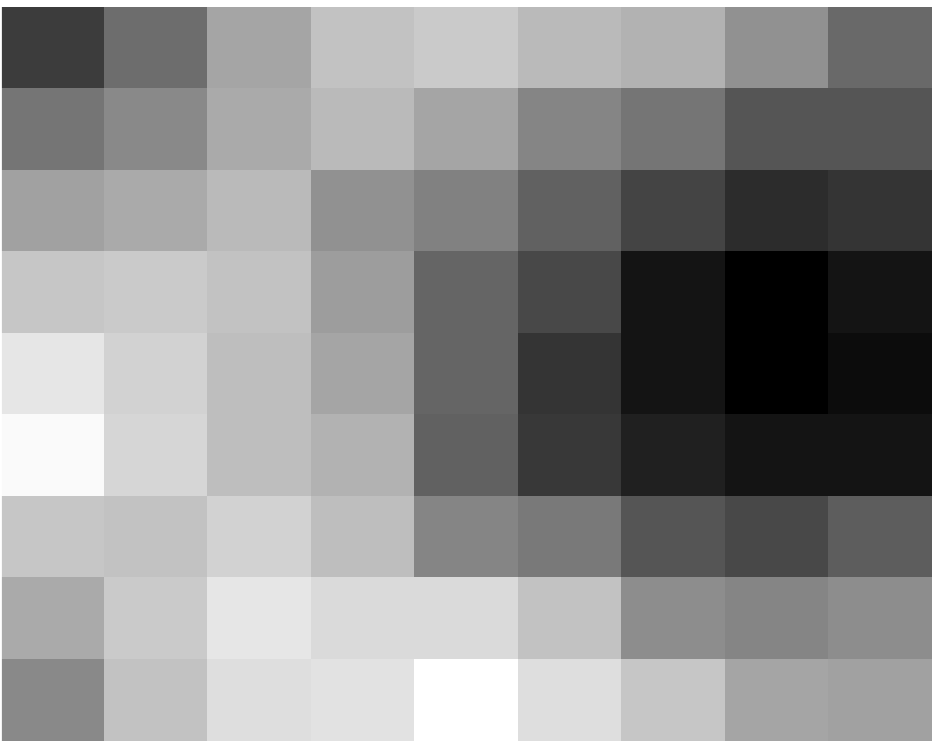} & 
\includegraphics[width=0.07\linewidth]{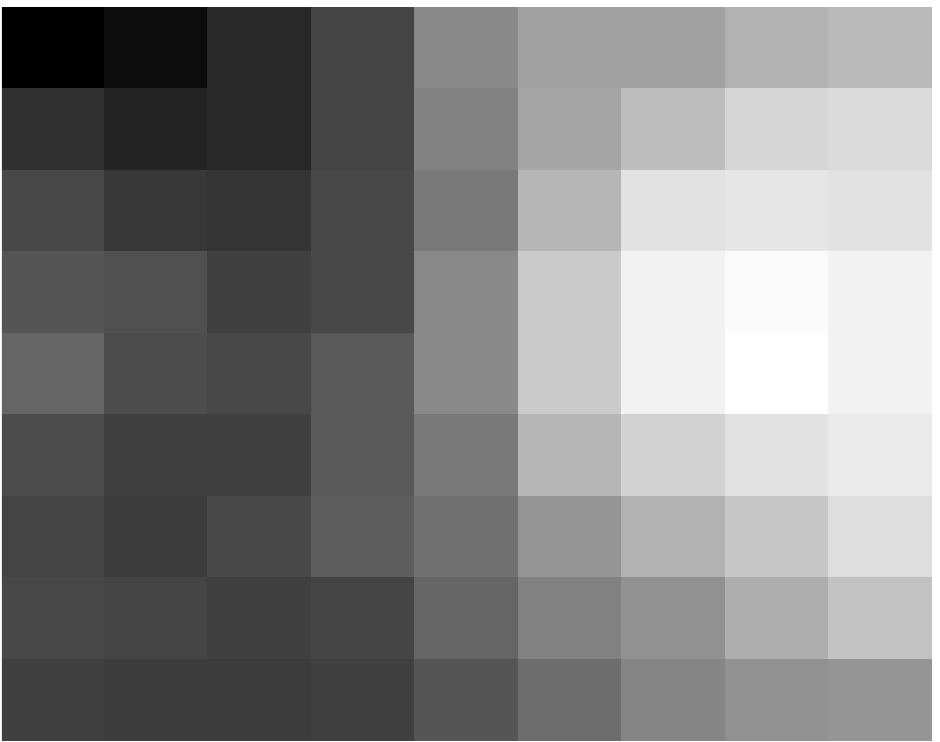} &
\includegraphics[width=0.07\linewidth]{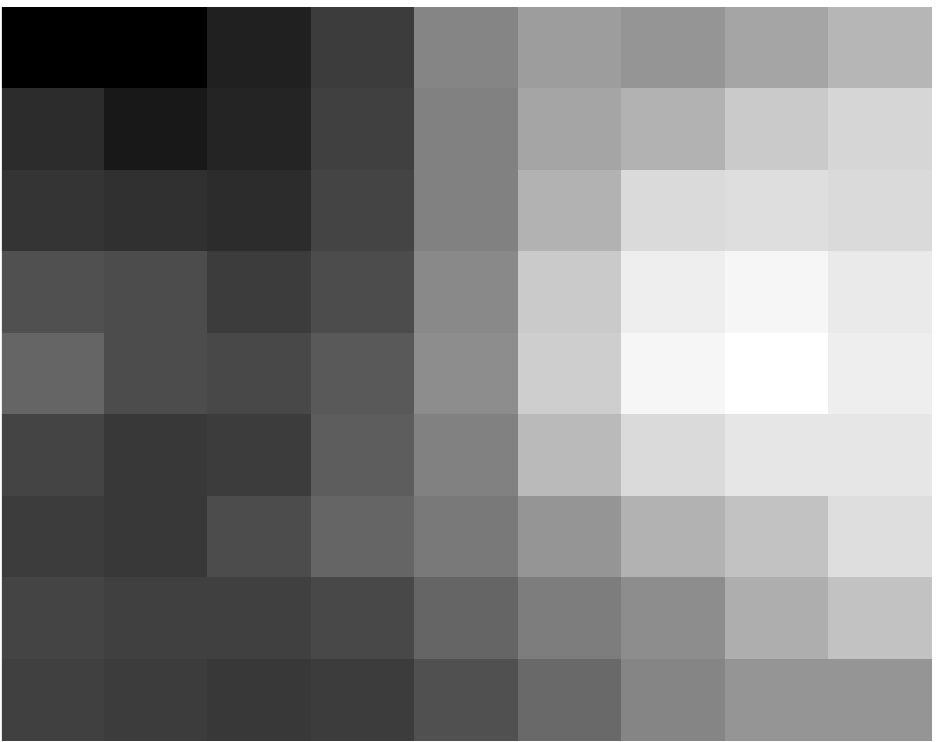} \\\hline

\multicolumn{1}{|c}{} &\multicolumn{8}{c|}{dataset: CIFAR-10 ; dictionary size: 1600}\\\hline

smooth \& batches &
\includegraphics[width=0.07\linewidth]{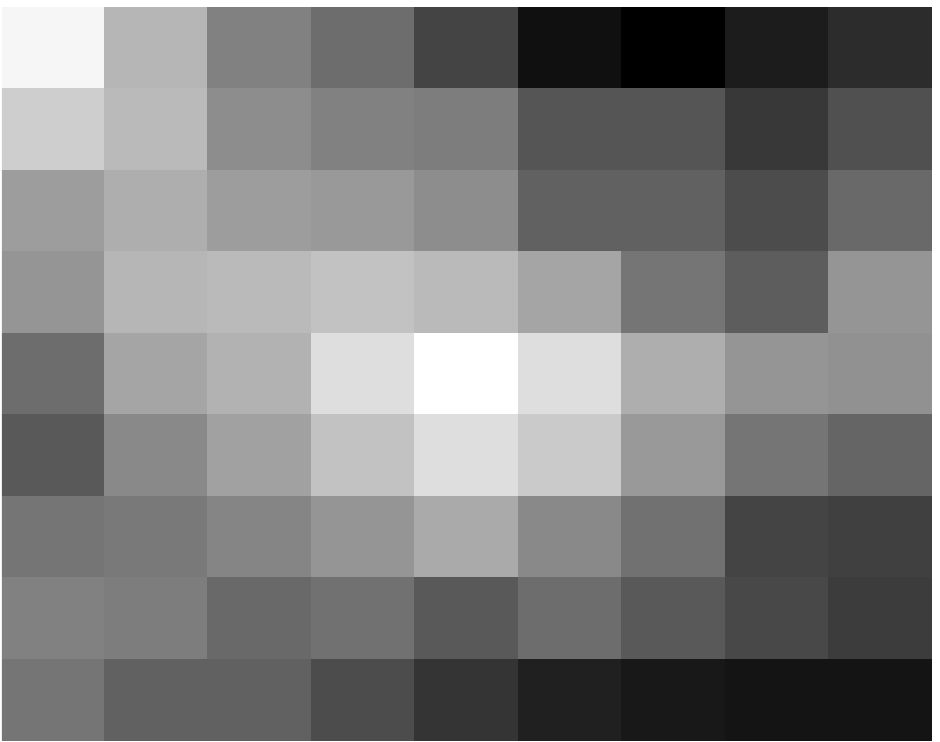} &
\includegraphics[width=0.07\linewidth]{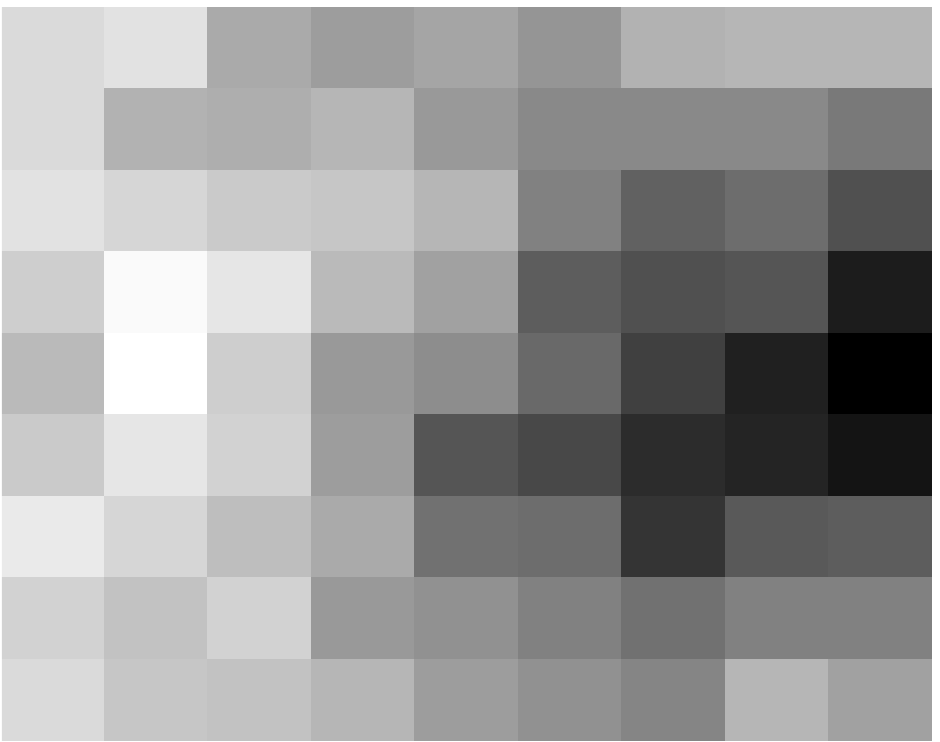} & 
\includegraphics[width=0.07\linewidth]{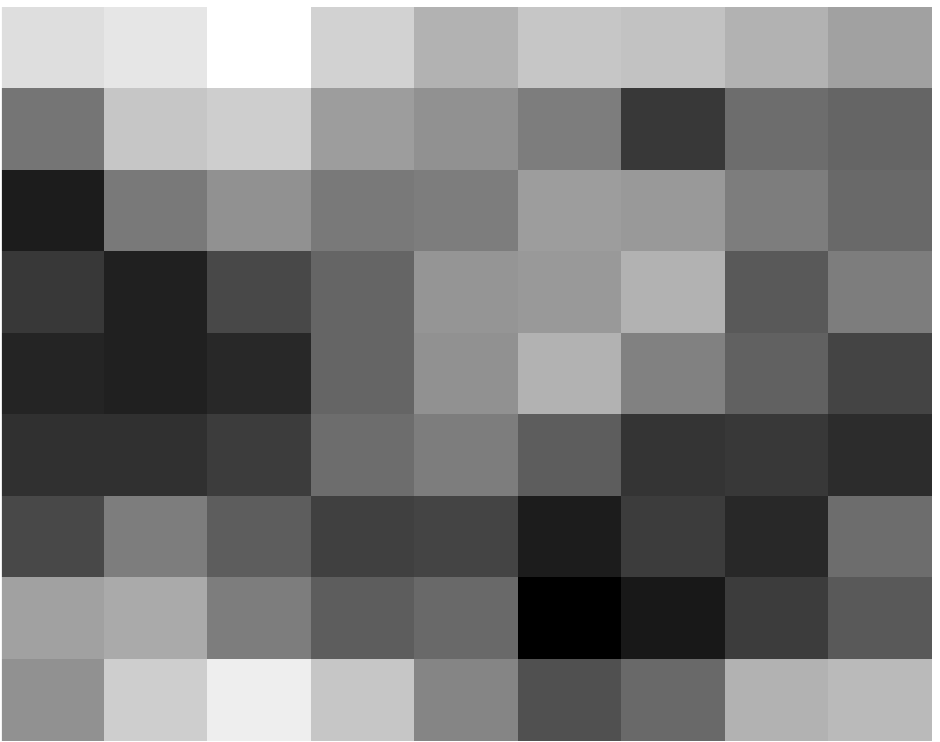} &
\includegraphics[width=0.07\linewidth]{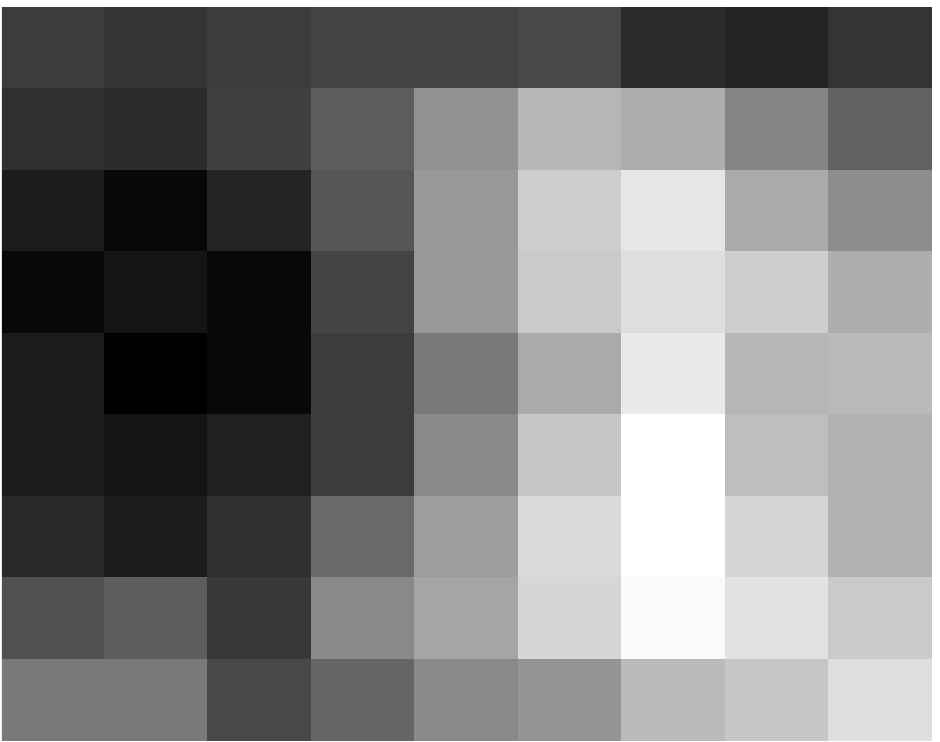} &
\includegraphics[width=0.07\linewidth]{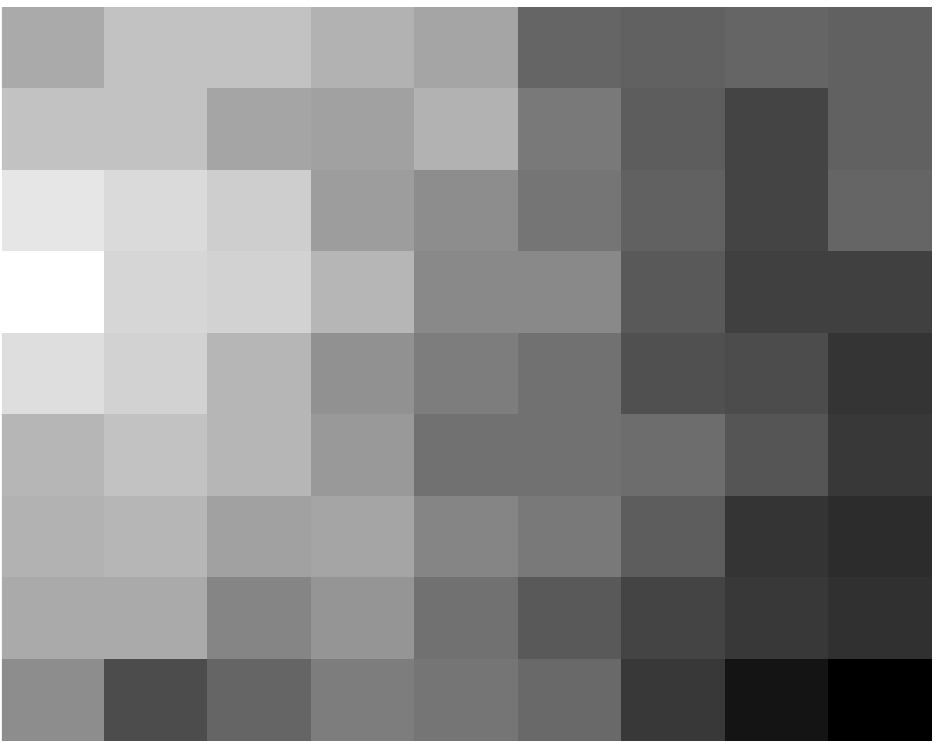} &
\includegraphics[width=0.07\linewidth]{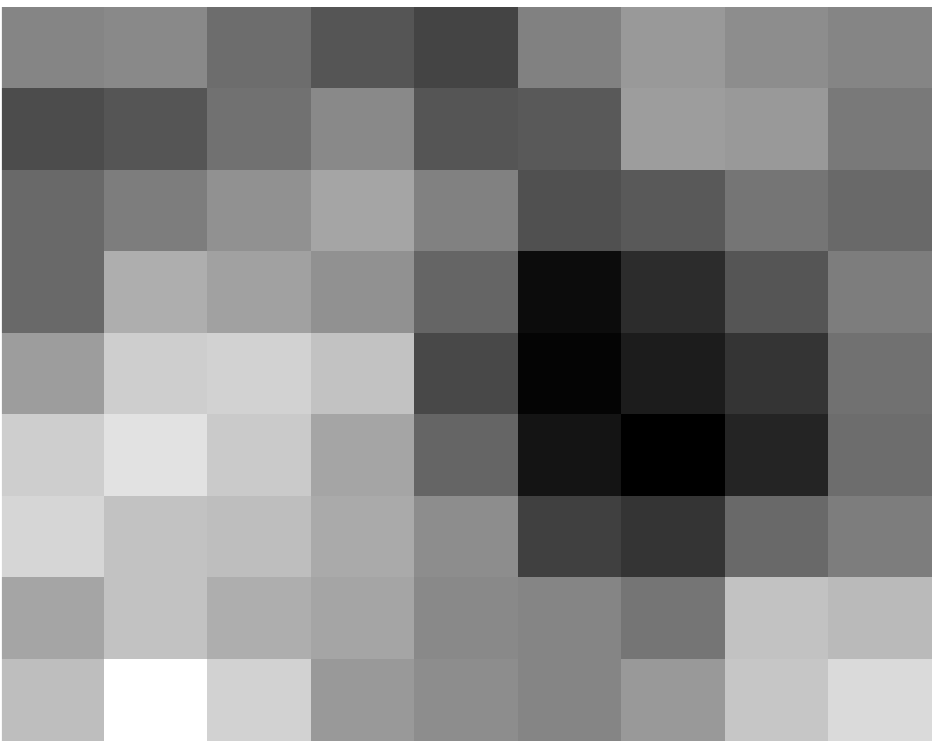} & 
\includegraphics[width=0.07\linewidth]{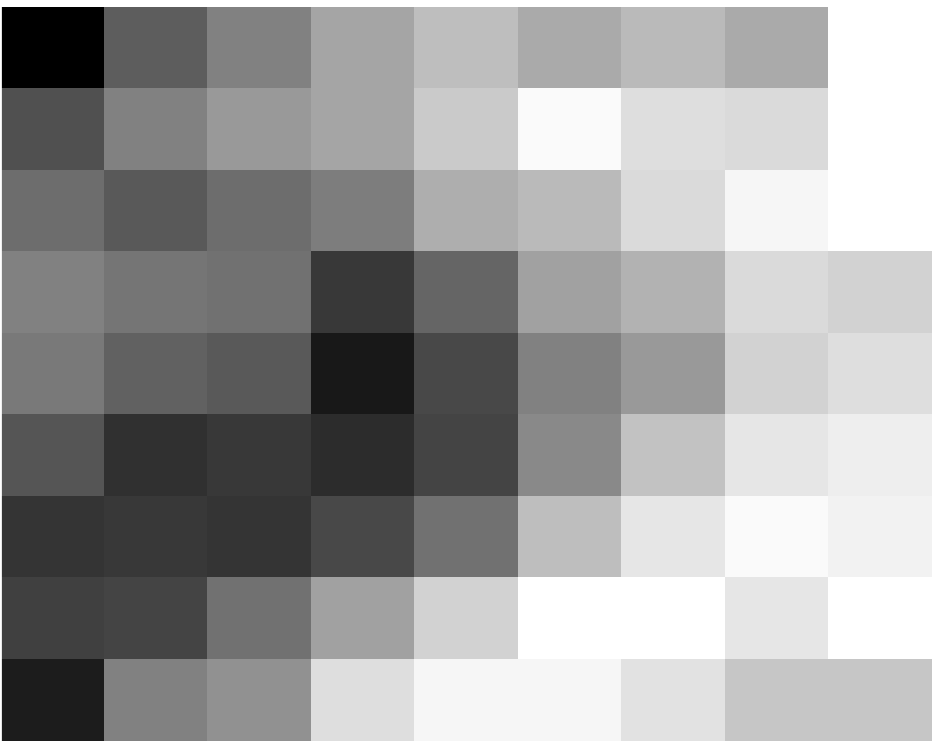} &
\includegraphics[width=0.07\linewidth]{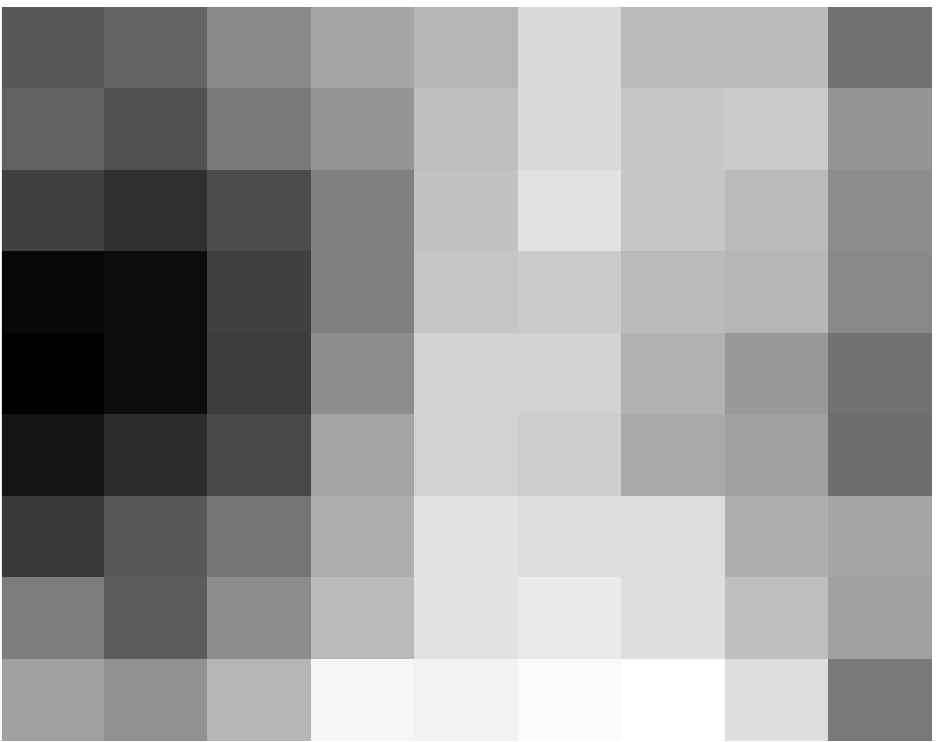} \\\hline

\multicolumn{1}{|c}{} & \multicolumn{8}{c|}{dataset: CIFAR-100 ; dictionary size: 1600}\\\hline

smooth \& batches & 
\includegraphics[width=0.07\linewidth]{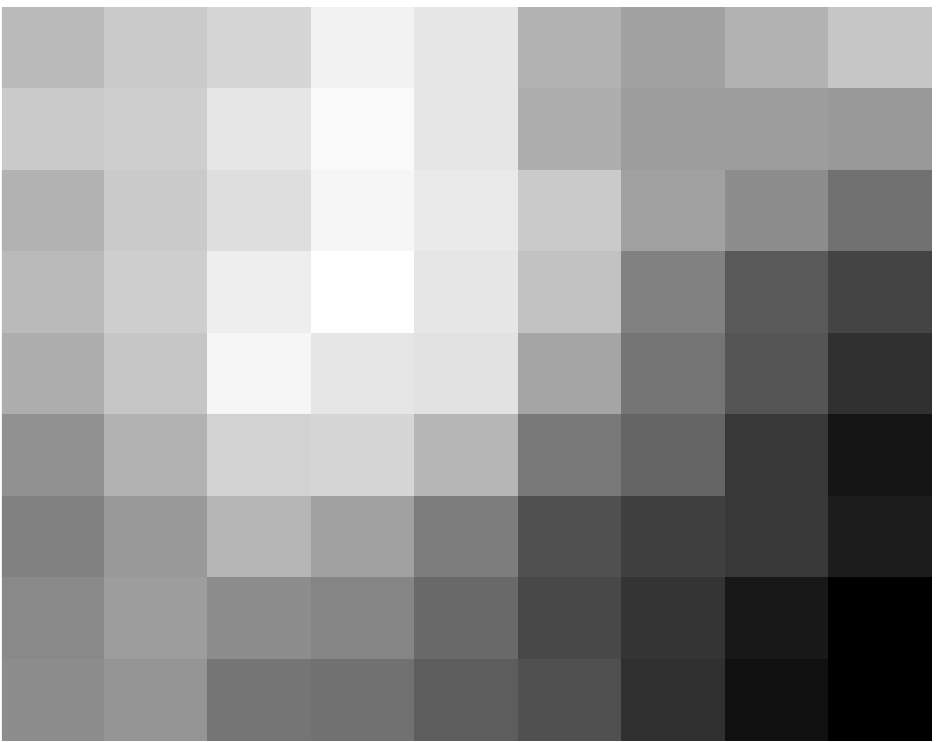} &
\includegraphics[width=0.07\linewidth]{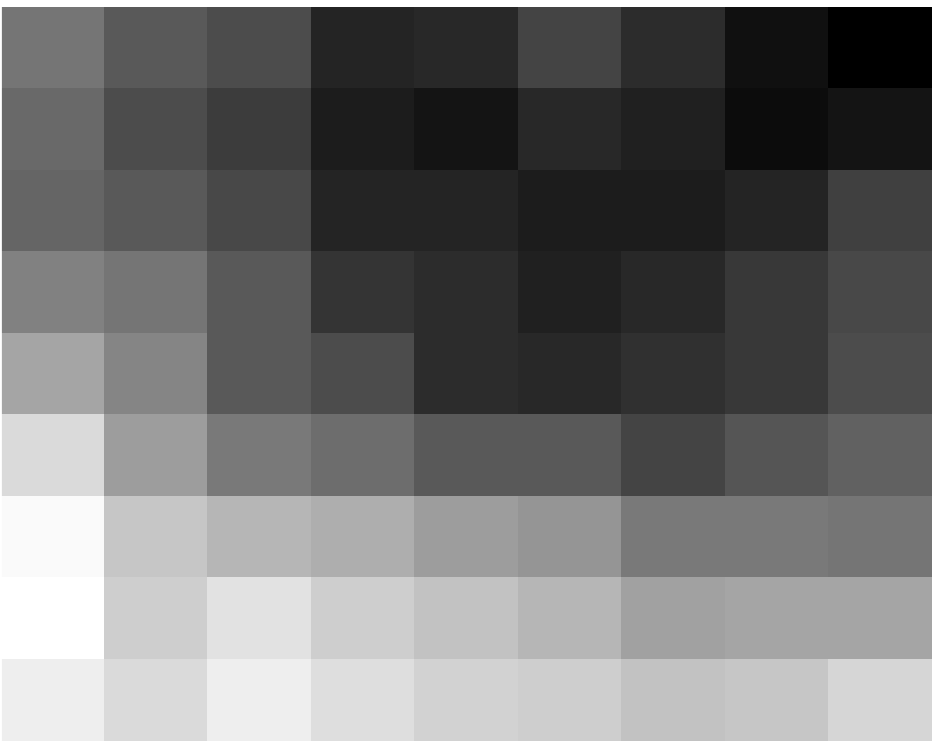} & 
\includegraphics[width=0.07\linewidth]{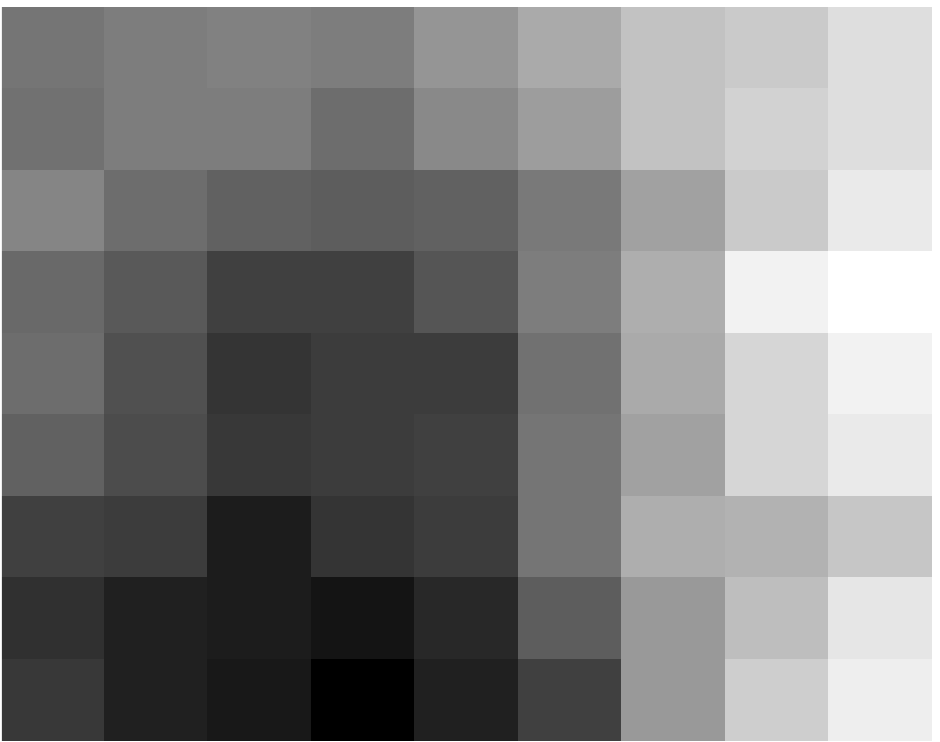} &
\includegraphics[width=0.07\linewidth]{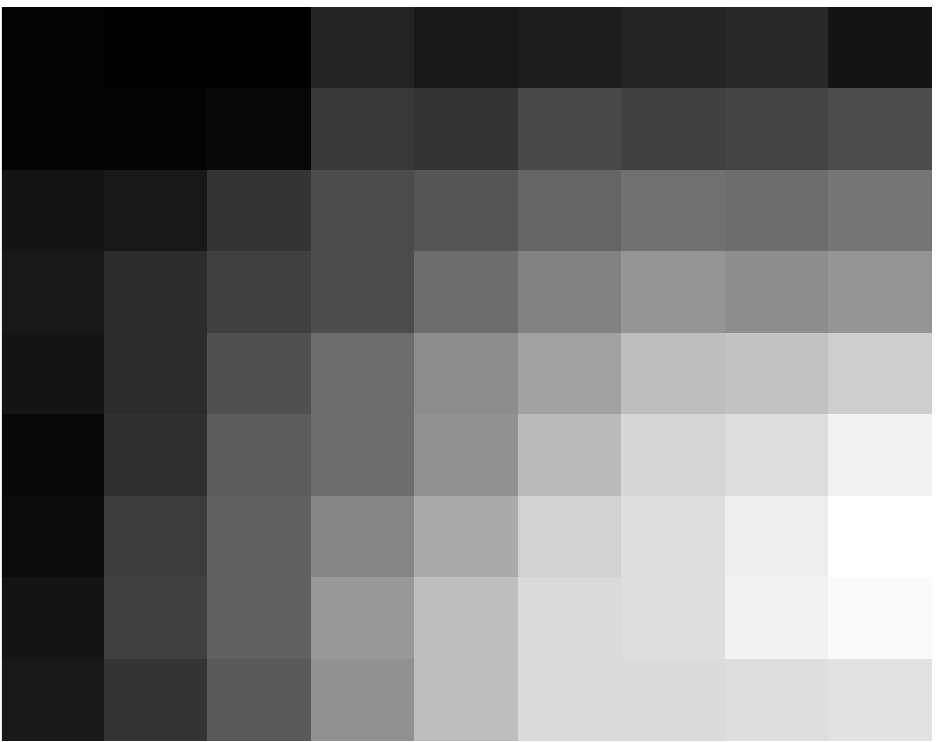} &
\includegraphics[width=0.07\linewidth]{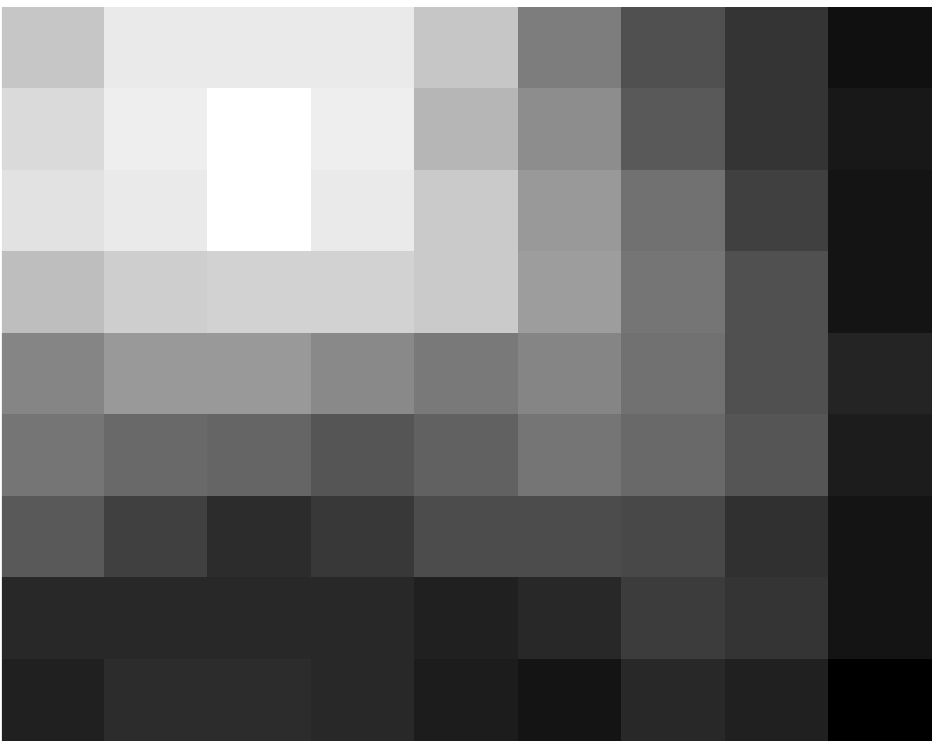} &
\includegraphics[width=0.07\linewidth]{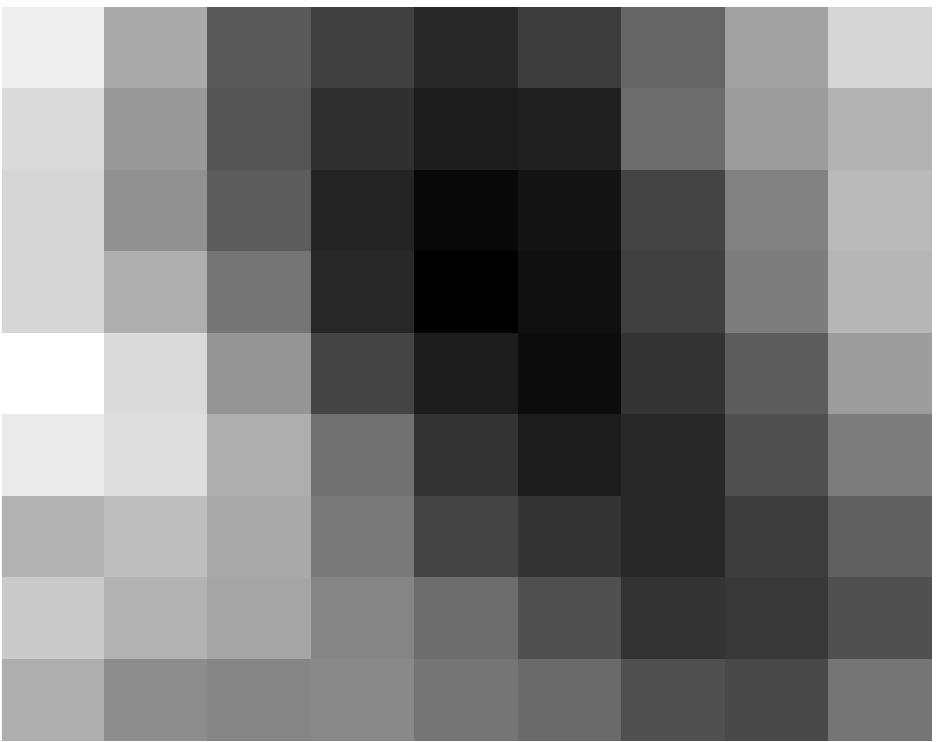} & 
\includegraphics[width=0.07\linewidth]{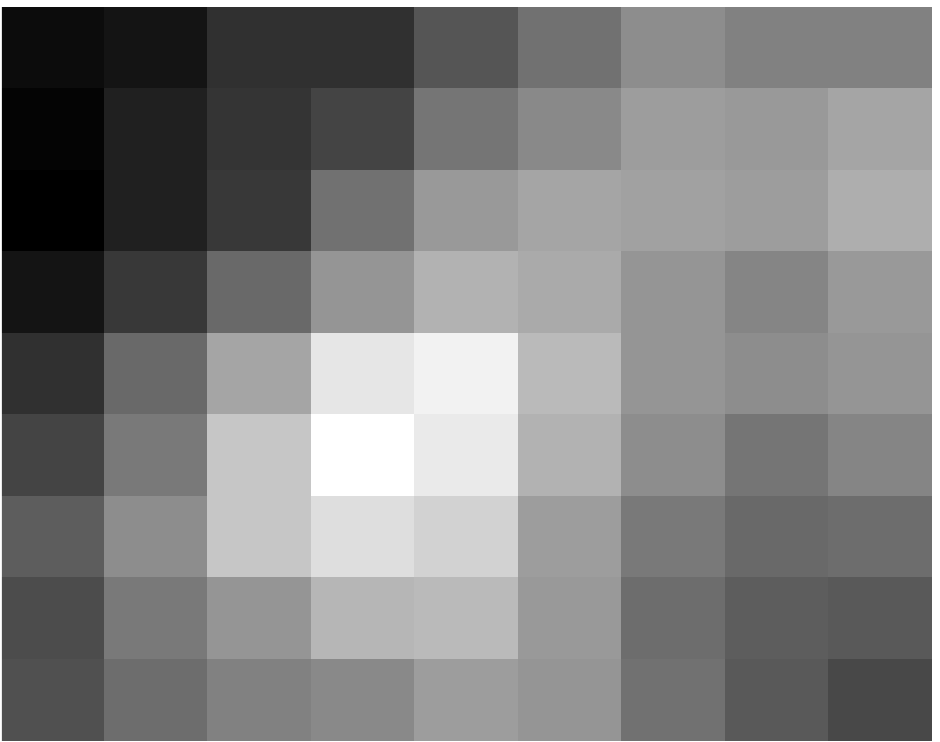} &
\includegraphics[width=0.07\linewidth]{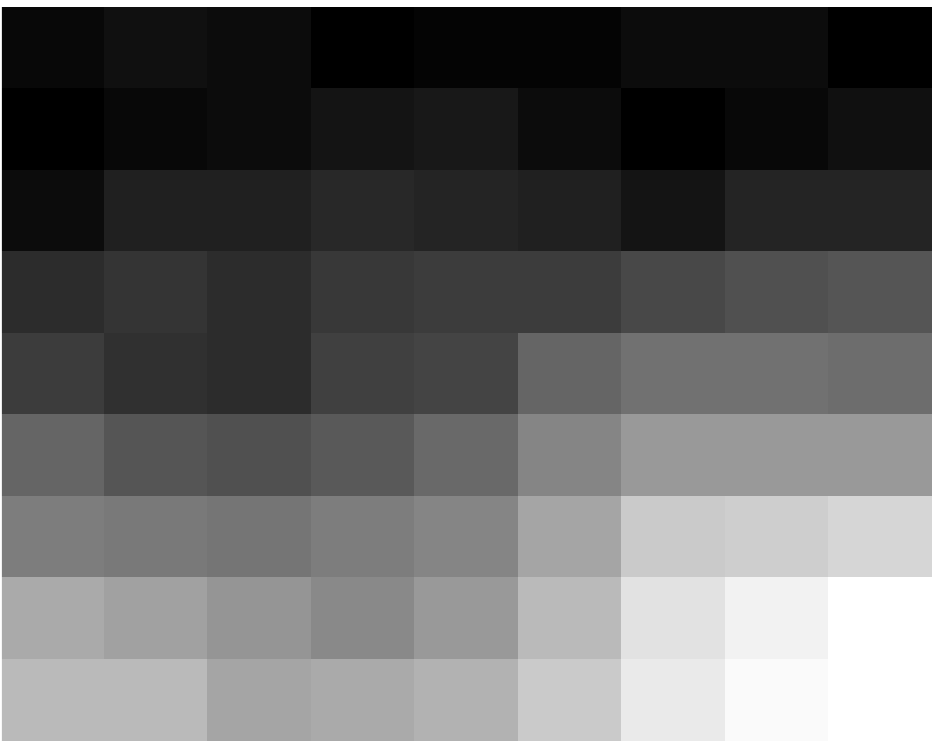} \\\hline

\end{tabular}
\caption{Visualization of different pooling strategies obtained for different regularizations, datasets and dictionary size. Every column shows the regions from two different coordinates of the codes. First row presents the initial configuration also used in standard hand-crafted pooling methods. Brighter regions denote larger weights.}
\label{fig:pooling_regions_viz}
\end{table*}

\subsection{Investigation of the regularization terms}
Our model (Eq. \ref{eq:learnable_pooling_regions_objective}) comes with two regularization terms associated with the pooling weights, each imposing different assumptions on the pooling regions. Hence, it is interesting to investigate their role in the classification task by considering all possible subsets of $\left\{\text{l2}, \text{smooth}\right\}$, where ``l2'' and ``smooth'' refer to $\twoNorm{\bs{W}}^2$ and 
$\left(\twoNorm{\nabla_x \bs{W}}^2 + 	\twoNorm{\nabla_y \bs{W}}^2\right)$ respectively.  

Table \ref{tab:regularization_experiments} shows our results on CIFAR-10. 
We choose a dictionary size of 200 for these experiments, so that we can evaluate different regularization terms without any approximations. We conclude that the spatial smoothness regularization term is crucial to achieve a good predictive performance of our method whereas the l2-norm term can be left out, and thus also reducing the number of hyper-parameters. 
Based on the cross-validation results (second column of Table \ref{tab:regularization_experiments}), we select this setting for further experiments.

\begin{table}[H]
\centering
\begin{tabular}{| l | c | r |}
	\hline
	Regularization &  CV Acc. & Test Acc. \\
	\hline 
 	free & $68.48\%$ & $69.59\%$ \\
	l2  & $67.86\%$ & $68.39\%$\\
	smooth & $73.36\%$ & $73.96\%$ \\
	l2 + smooth & $70.42\%$ & $ 70.32\%$ \\
	\hline
\end{tabular}
\caption{We investigate the impact of the regularization terms on the CIFAR-10 dataset with dictionary size equals to $200$. Term ``free'' denotes the objective function without the l2-norm and smoothness regularization terms. The cross-validation accuracy and test accuracy are shown.}
\label{tab:regularization_experiments}
\end{table}

\subsection{Experiments on the CIFAR-100 dataset}
Although the main body of work is conducted on the CIFAR-10 dataset, we also investigate how the model performs on the much more demanding CIFAR-100 dataset with $100$ classes.
Our model with the spatial smoothness regularization term on the $40$ dimensional batches achieves $56.29\%$ accuracy. To our best knowledge, this result consitutes the state-of-the-art performance on this dataset, outperforming \citet{yangqing11nips} by $1.41\%$, and the baseline by $4.63\%$. Using different architecture \citet{Goodfellow_maxout_2013} has achieved accuracy $61.43\%$.

\begin{table}[H]
\centering
\begin{tabular}{| l | c | c | r |}
	\hline
	Method & Dict. size & Features & Acc. \\
	\hline 
	Jia & $1600$ & $6400$ & $54.88\%$ \\
\hline
 	Coates & $1600$ & $6400$ & $51.66\%$ \\	
	Our (batches) & $1600$ & $6400$ & $56.29\%$\\
	\hline
\end{tabular}
\caption{The classification accuracy on CIFAR-100, where our method is compared against the 
\citet{coates2011importance} (we downloaded the framework from https://sites.google.com/site/kmeanslearning, we also use $5$-fold cross-validation to choose hyper-parameter $C$) and \citet{yangqing11nips} (here we refer to the NIPS 2011 workshop paper).}
\label{tab:coates_receptive_field_pooling_experiments}
\end{table} 

\subsection{Transfer of the pooling regions between datasets}
Beyond the standard classification task, we also examine if the learnt pooling regions are transferrable between datasets. In this scenario the pooling regions are first trained on the source dataset and then used on the target dataset to train a new classifier. We use dictionary of $1600$ with $40$-dimensional batches. Our results (Table \ref{tab:transfer_datasets_experiments}) suggest that the learnt pooling regions are indeed transferable between both datasets. While we observe a decrease in performance when learning the pooling strategy on the less diverse CIFAR-10 dataset, we do see improvements for learning on the richer CIFAR-100 dataset. We arrive at a test accuracy of $80.35\%$ which is an additional improvement of $0.75\%$ and $0.18\%$ over our best results (batch-based approximation) and \citet{yangqing11nips} respectively. 

\begin{table}[H]
\centering
\begin{tabular}{| l | c | r |}
	\hline
	Source & Target & Accuracy \\
	\hline 
 	CIFAR-10 & CIFAR-100  & $52.86\%$ \\
	CIFAR-100 & CIFAR-10  & $80.35\%$ \\
	\hline
\end{tabular}
\caption{We train the pooling regions on the 'Source' dataset. Next, we use such regions to train the classifier on the 'Target' dataset where the test accuracy is reported. 
}
\label{tab:transfer_datasets_experiments}
\end{table}

\subsection{Visualization and analysis of pooling strategies}
\label{subsection:visualization_of_pooling_regions}
Table \ref{fig:pooling_regions_viz} visualizes different pooling strategies investigated in this paper.
The first row shows the widely used rectangular spatial division of the image.
The other visualizations correspond to pooling weights discovered by our model using different regularization terms, datasets and dictionary size.
The second row shows the results on CIFAR-10 with the ``l2'' regularization term. The pooling is most distinct from the other results, as it learns highly localized weights. This pooling strategy has also performed the worst in our investigation (Table \ref{tab:regularization_experiments}).

The ''smooth'' pooling performs the best. Visualization shows that weights are localized but vary smoothly over the image. The weights expose a bias towards initialization shown in the first row. All methods with the spatial smoothness regularization tend to focus on similar parts of the image, however ``l2 \& smooth'' is more conservative in spreading out the weights.

The last two rows show weights trained using our approximation by batches.
From visual inspection, they show a similar level of localization and smoothness to the regions obtained without approximation. This further supports the use of our approximation into independent batches. 

\section{Conclusion}
\label{section:conclusion}
In this paper we propose a flexible parameterization of the pooling operator which can be trained jointly with the classifier.
In this manner, we study the effect of different regularizers on the pooling regions as well as the overall system. To be able to train the large set of parameters we propose approximations to our model allowing efficient and parallel training without loss of accuracy.

Our experiments show there is a room to improve the classification accuracy by advancing the spatial pooling stage.
The presented method outperforms a popular hand-crafted pooling based method and previous approaches to learn pooling strategies. While our improvements are consistent over the whole range of dictionary sizes that we have investigated, the margin is most impressive for small codes where we observe improvements up to $10\%$ compared to the baseline of Coates. Finally, our method achieves an accuracy of $56.29\%$ on CIFAR-100, which is to the best of our knowledge the new state-of-the-art on this dataset. 

As we believe that our method is a good framework for further investigations of different pooling strategies and in order to speed-up progress on the pooling stage
we will make our code publicly available at time of publication.

\bibliographystyle{unsrtnat}
\small{
\bibliography{egbib_short}
}
\end{document}